\documentclass{article}

\usepackage[utf8]{inputenc} %
\usepackage[margin=1in]{geometry}
\usepackage[authoryear]{natbib}
\usepackage[T1]{fontenc}    %
\usepackage{hyperref}       %
\usepackage{url}            %
\usepackage{booktabs}       %
\usepackage{amsfonts}       %
\usepackage{nicefrac}       %
\usepackage{microtype}      %
\usepackage{xcolor}         %
\usepackage{graphicx}
\usepackage{xcolor}
\usepackage{listings}
\usepackage{epigraph}
\usepackage{subcaption}

\usepackage{wrapfig} 

\definecolor{lightgray}{gray}{0.95}

\newcounter{taskcounter}

\usepackage{nicefrac}
\usepackage{hyperref}
\hypersetup{
    colorlinks=true,
    linkcolor=blue,
    filecolor=magenta,      
    urlcolor=cyan
}
\usepackage{mathtools}
\usepackage{amsmath,amsfonts}

\usepackage{amssymb}
\usepackage[ruled,vlined]{algorithm2e}
\usepackage{algorithmic}
\usepackage{makecell}
\usepackage{diagbox}
\usepackage{enumitem}
\usepackage{multirow}
\usepackage{textgreek}
\usepackage{comment}

\usepackage{caption}
\usepackage{subcaption}

\usepackage{amsthm}

\usepackage[nameinlink,capitalize]{cleveref}

\crefname{section}{\S}{\S}
\Crefname{section}{\S}{\S}
\crefname{appendix}{App.}{Apps.}
\Crefname{appendix}{App.}{Apps.}
\crefname{theorem}{Thm.}{Thms.}
\Crefname{theorem}{Thm.}{Thms.}
\crefname{proposition}{Prop.}{Props.}
\Crefname{proposition}{Prop.}{Props.}
\crefname{algorithm}{Alg.}{Algs.}
\Crefname{algorithm}{Alg.}{Algs.}
\crefname{assumption}{Asm.}{Asms.}
\Crefname{assumption}{Asm.}{Asms.}
\crefname{mechanism}{Mech.}{Mechs.}
\Crefname{mechanism}{Mech.}{Mechs.}

\usepackage{mfirstuc}
\usepackage[T1]{fontenc}

\newcounter{packednmbr}

\newcommand{\rebuttal}[1]{#1}

\newcommand{\base}{\mathcal{D}}
\newcommand{\synthetic}{\mathcal{D}'}
\newcommand{\name}{Struct-Bench\xspace}

\newcommand{\calK}{\mathcal{K}}

\newcommand{\calM}{\mathcal{M}}

\newcommand{\calS}{\mathcal{S}}

\newcommand{\bra}[1]{\left( #1 \right)}

\newcommand{\brc}[1]{\left\{ #1 \right\}}

\newcommand{\probnotation}{\mathbb{P}}
\newcommand{\probof}[1]{\probnotation\bra{#1}}

\newcommand{\functionrange}[1]{range\bra{#1}}

\newcommand{\dpmechanismnotation}{\calM}
\newcommand{\dpmechanismof}[1]{\dpmechanismnotation\bra{#1}}

\newcommand{\sampleattribute}{a}
\newcommand{\objectattribute}{v}
\newcommand{\dataset}{\mathcal D}
\newcommand{\sample}{D}
\newcommand{\object}{O}

\newcommand{\cfg}{CFG-PR}
\newcommand{\kns}{KND}
\newcommand{\knd}{AM}
\newcommand{\precision}{KNN-Precision}
\newcommand{\recall}{KNN-Recall}

\newcommand{\PE}{PE}
\newcommand{\DPFT}{DP-FT}
\newcommand{\IF}{IF}

\newcommand{\FT}{FT}
\newcommand{\downstream}{Acc}

\newcommand{\bench}{Struct-Bench}

\crefformat{section}{§#2#1#3}

\title{
Struct-Bench: A Benchmark for Differentially Private Structured Text Generation
}

\author{
  Shuaiqi Wang\thanks{ These authors contributed equally to this work.}\ \thanks{This work was partially done while an intern at Microsoft.} \\
  \small{Carnegie Mellon University}\\
  \small{\texttt{shuaiqiw@andrew.cmu.edu}}
  \and
  Vikas Raunak\footnotemark[1]\ \thanks{Now at Google DeepMind.} \\
  \small{Microsoft Corporation}\\
  \small{\texttt{viraunak@microsoft.com}}
  \and
  Arturs Backurs \\
  \small{Microsoft Research}\\
  \small{\texttt{arturs.backurs@microsoft.com}}
  \and
  Victor Reis \\
  \small{Microsoft Research}\\
  \small{\texttt{victorol@microsoft.com}}
  \and
  Pei Zhou\thanks{Primary internship mentors: Pei Zhou, Zinan Lin.} \\
  \small{Microsoft Corporation}\\
  \small{\texttt{pei.zhou@microsoft.com}}
  \and
  Sihao Chen \\
  \small{Microsoft Corporation}\\
  \small{\texttt{sihaochen@microsoft.com}}
  \and
  Longqi Yang \\
  \small{Microsoft Corporation}\\
  \small{\texttt{Longqi.Yang@microsoft.com}}
  \and
  Zinan Lin\footnotemark[4] \\
  \small{Microsoft Research}\\
  \small{\texttt{zinanlin@microsoft.com}}
  \and
  Sergey Yekhanin \\
  \small{Microsoft Research}\\
  \small{\texttt{yekhanin@microsoft.com}}
  \and
  Giulia Fanti \\
  \small{Carnegie Mellon University}\\
  \small{\texttt{gfanti@andrew.cmu.edu}}
  }

\date{}

\begin{document}

\maketitle

\begin{abstract}
Differentially private (DP) synthetic data generation  is a promising technique for utilizing private datasets that otherwise cannot be exposed for model training or other analytics. While much research literature has focused on generating private unstructured text and image data, in enterprise settings, \emph{structured data} (e.g., tabular) is more common, often including natural language fields or components. 
Existing synthetic data evaluation techniques (e.g., FID) struggle to capture the structural properties and correlations of such datasets.
In this work, we propose Struct-Bench, a framework and benchmark 
for evaluating synthetic datasets derived from  structured datasets that contain natural language data. 
The Struct-Bench framework requires users to provide a representation of their dataset structure as a Context-Free Grammar (CFG). Our benchmark comprises 5 real-world and 2 synthetically generated datasets, each annotated with
CFGs.
We show that these datasets demonstrably present a great challenge even for state-of-the-art DP synthetic data generation methods. 
Struct-Bench also includes reference implementations of different metrics and a leaderboard, thereby providing researchers a standardized evaluation platform to benchmark and investigate privacy-preserving synthetic data generation methods. Further, we also present a case study showing how to use Struct-Bench to improve the synthetic data quality of Private Evolution (PE) on structured data.
The benchmark and the leaderboard have been publicly made available at \url{https://struct-bench.github.io}.
\end{abstract}
\section{Introduction}

Enterprise settings often feature datasets  that include both \emph{structured} relationships between fields or objects, and fields that contain \emph{natural language data}.
For example, consider a dataset of user queries to a search engine, and the corresponding results. The dataset is structured in a question-and-response structure, and both the query and the response contain natural language. 
{Another example consists of patients' medical records, which can include multiple events over time, visits with different providers for different ailments, and associated (natural language) doctors' notes.}
Such datasets are valuable for downstream use cases (e.g.,  training predictive models, understanding preferences), but they cannot always be used directly, due to privacy or data use restrictions. 

Differentially private (DP) synthetic data generation is an increasingly important technique for making use of such sensitive datasets \cite{zhang2021privsyn,rosenblatt2020differentially,xie2024differentially}. 
Evaluating the quality of such synthetic data (private or not) is an active research area, with many proposals tailored to unstructured data like images or text \cite{torkzadehmahani2019dp,lin2023differentially,xie2024differentially,gong2025dpimagebench} and tabular data \cite{mckenna2022aim,tran2024differentially,zhang2021privsyn}.
Notably, existing synthetic data evaluation frameworks do not naturally capture the salient properties of datasets that feature both general structural properties \emph{and} natural language elements. Evaluation metrics for unstructured data, like Fréchet Inception Distance (FID) \cite{heusel2017gans}, or precision and recall \cite{kynkaanniemi2019improved,sajjadi2018assessing},  do not capture the structural properties of data. {In fact, as we will show in our results, synthetic data that completely fails to capture structural constraints can still achieve high precision and recall scores. For example, the ShareGPT dataset \cite{anon_sharegpt_vicuna_unfiltered_commit_bcd32a7}, which consists of multi-round conversations between a human and an AI agent, requires the format tokens `HUMAN:' before queries and `GPT:' before responses. Consider the synthetic sample: "How are you? I’m doing well." Although semantically reasonable—and therefore scoring high on precision—it violates the required format. This underscores the importance of structure-aware evaluation.} 
On the other hand, evaluation frameworks for tabular data are designed primarily for datasets with categorical or numeric fields \cite{hernadez2023synthetic,yang2024structured,chundawat2022universal}, and do not naturally extend to natural language fields. For instance, these frameworks often compare $k$-way marginal distributions in the real and synthetic data, which is not meaningful for natural language.

\begin{figure}[t]
\vspace{-1mm}
    \centering
    \begin{subfigure}{0.6\textwidth}
        \centering
        \includegraphics[width=\textwidth]{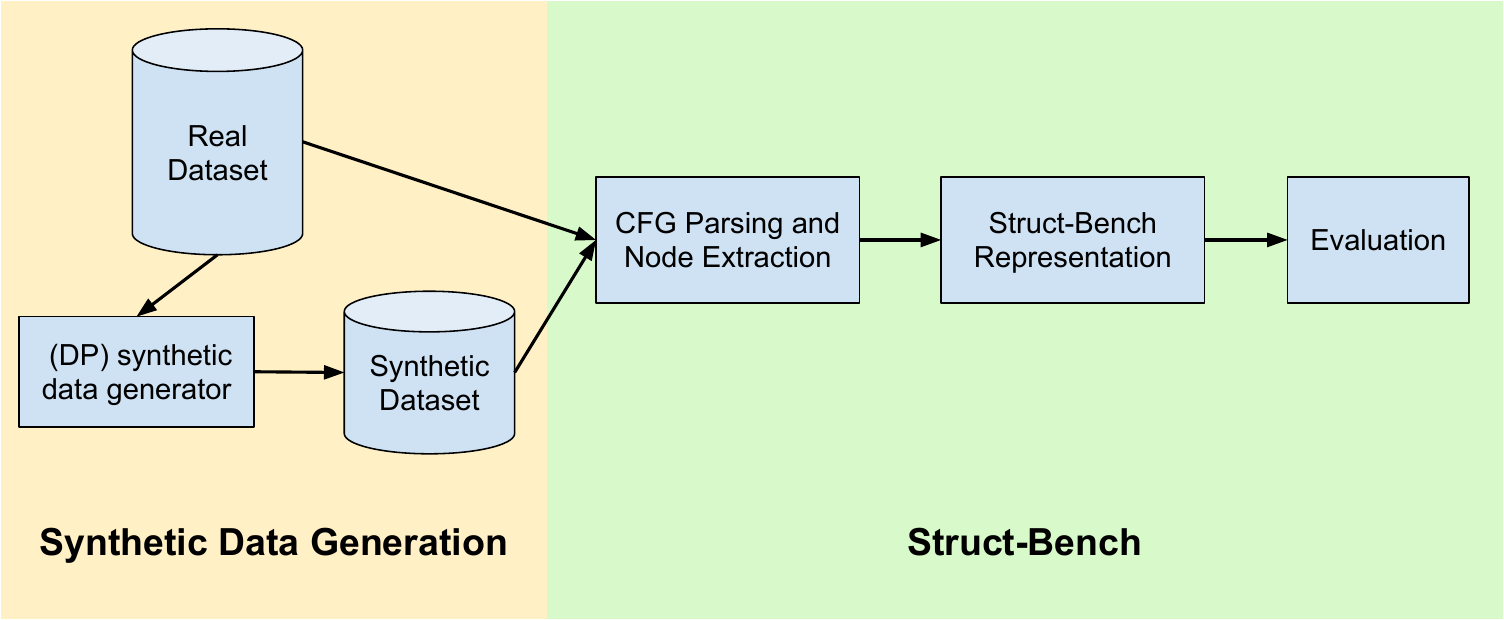}
        \caption{\bench{} evaluation pipeline. The synthetic dataset can be generated via DP generation methods with private access to the private dataset. \bench{} evaluates a synthetic dataset by parsing samples and extracting nodes and their attributes using context-free grammar (CFG) with full access to both private and synthetic datasets. }
        \label{fig:dataset_level}
    \end{subfigure}
    \hfill
    \begin{subfigure}{0.39\textwidth}
        \centering
        \includegraphics[width=\textwidth]{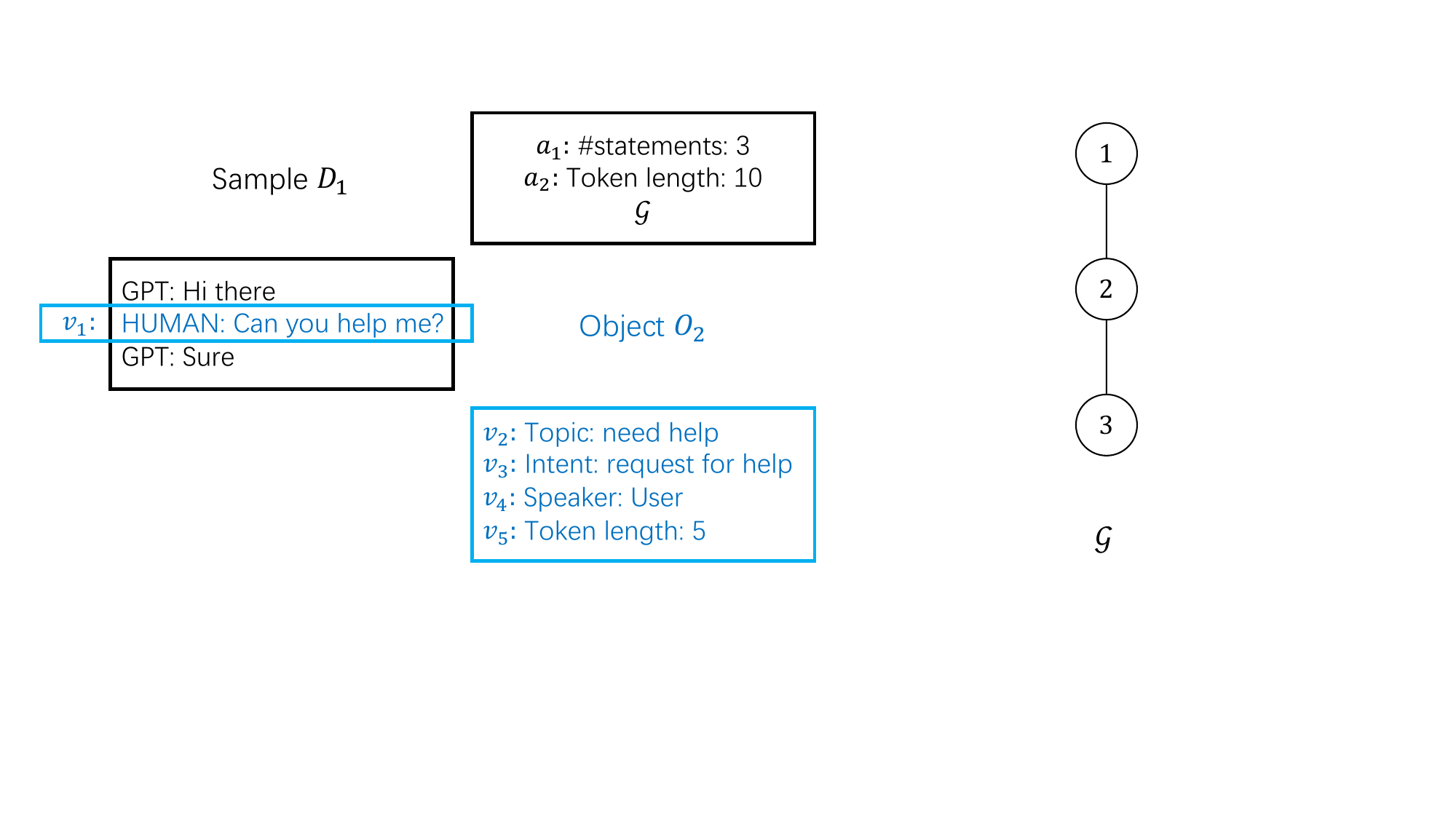}
        \caption{Sample level view of the \bench{}. {As an example, a} multi-round conversation is parsed by CFG into several nodes with types Query and Response. \bench{} extracts its sample-level and node-level attributes for evaluation.%
        }
        \label{fig:sample_level}
    \end{subfigure}
    \caption{Dataset level and sample level views into the Struct-Bench framework. %
    }
    \label{fig:two-subfigs}
    \vspace{-2mm}
\end{figure}

In this work, drawing inspiration from Natural Language Generation (NLG) benchmarks such as GEM \cite{gem}, we present \textit{Struct-Bench}: {a composite benchmark and an automatic evaluation protocol} that measures the  quality of structured, natural language-based synthetic data relative to their corresponding real (possibly private) datasets. In the Struct-Bench framework, a user first selects one or more real datasets for which they want to evaluate 
a corresponding synthetic dataset. For each dataset, the user provides a set of production rules under the generative grammar formalism (i.e., a context-free grammar \cite{hopcroft2001introduction}); %
the user then selects \textit{key nodes}, which will be programmatically extracted from the parse tree of dataset samples; key nodes are used to measure important correlations and properties of the synthetic data. 
Based on this dataset representation, Struct-Bench measures an array of syntactic (i.e., structural) and semantic properties of the synthetic dataset. %
We illustrate the dataset level and sample level views into \bench{} in \cref{fig:two-subfigs}.%

Our contributions are as follows:

\begin{enumerate}
    \item %
    \textbf{Benchmark:} %
    We propose Struct-Bench, a novel evaluation framework and a benchmark to evaluate synthetic data quality relative to a real dataset, where the real dataset features complex inter-field structural relations, and at least some fields contain  natural language.
    A key observation is that many structural properties (and even semantic properties) can be modeled and evaluated with the help of a context-free grammar. %
    We also provide a public leaderboard and a reference, extensible implementation of the Struct-Bench framework, which can be used to benchmark new DP synthetic data generation methods on other datasets in a standardized manner.%
    \item %
    \textbf{Findings:} %
    We use Struct-Bench to benchmark and analyze state-of-the-art (SOTA) DP synthetic data generation techniques on %
    seven diverse datasets, including real-world textual and tabular datasets, as well as synthetic datasets with controllable data attributes. %
    Our main findings are (1) no single metric fully describes synthetic data quality; (2) %
    {none of the existing SOTA DP synthetic data generation techniques are able to reliably capture the structural properties of data without sacrificing semantic performance.}
    Finding (1) highlights the importance of using multiple metrics to evaluate synthetic data, which is a key contribution of our work, and (2) underscores the need for further research in synthetic structured data generation. %
    We also conduct a case study to show how to algorithmically improve on SOTA DP synthetic data generation methods, Private Evolution (PE) \cite{lin2023differentially,xie2024differentially,lin2025differentially}, using the insights from Struct-Bench. 
    These improvements achieve nearly 100\% compliance with dataset structural constraints, while also improving semantic and statistic metrics.  
\end{enumerate}

\section{Struct-Bench Framework and Evaluation Protocol}
\label{sec:framework}

We aim to design an evaluation framework that measures how closely a real, {private} dataset $\base$ matches a synthetic dataset $\synthetic$. The real dataset $\base$ features (1) an inherent \emph{structure}, and (2) \emph{natural language} fields. 
Our evaluation framework must therefore quantify how well $\synthetic$ has acquired the structure and content of the private dataset. 
As these are not well-defined quantities, we define a framework for representing a real dataset $\base$, as well as a suite of metrics to capture how well the synthetic dataset matches the syntax and semantics of the real dataset.

\vspace{-1mm}
\paragraph{Notation}
Consider a dataset $\mathcal{D}=(\sample_i)_{i=1}^{m}$ with $m$ samples. 
As a concrete running example, suppose $\dataset$ is the ShareGPT dataset \cite{anon_sharegpt_vicuna_unfiltered_commit_bcd32a7}, %
which contains multi-round conversations between a human and an AI agent. {We illustrate a sample of ShareGPT in \cref{app:sharegpt}.}
A dataset is a set of \emph{samples}; in ShareGPT,
each sample $\sample_i \in \base$ is one full conversation (e.g., a few iterations of conversation between the human and the agent).
Each sample $\sample_i$  contains a set of nodes $O_1, O_2, \ldots, O_{n_i}$ (the number of nodes can vary across samples). 
In the ShareGPT example, each node $O_j$ is one text snippet---either a single query from the human or a response from the AI agent, %
as illustrated in \cref{fig:sample_level}. %

Samples and nodes can have \emph{attributes}, which are either numeric or categorical. 
Each sample $\sample_i$ is associated with \emph{sample-level attributes} $\sampleattribute_1, \ldots, \sampleattribute_m$. 
These can be any derived property of the whole samples, such as token length of the conversation in ShareGPT.
A  node $O_j$ can also have \emph{node-level attributes}  $v_{1}, \ldots, v_{n_j}$. 
In ShareGPT, attributes could include the token length of a node, the identity of the speaker (agent or human), and the topic of the query or response {(obtained through human or LLM-based labeling)}.

\vspace{-1mm}
\paragraph{Dataset Representation}
The nodes of a dataset may satisfy complex structural relations. For example, in ShareGPT, each node can only have \texttt{human} or \texttt{agent} as its \texttt{speaker} attribute, and successive nodes should always alternate speaker between \texttt{human} and \texttt{agent}.
This information is considered public (to the synthetic data holder); for example, if an enterprise is training a DP synthetic dataset to model a private dataset of search engine queries, the enterprise is likely to know the \emph{schema} of the data, even if they do not know the contents.
Such relations are captured in Struct-Bench by a \emph{context-free grammar (CFG)}, which specifies different categories of nodes and the relations between them. As an example, we provide the CFG of ShareGPT in \cref{app:sharegpt}.
To add a new dataset to Struct-Bench, a user must write structural dependencies  that should be enforced in the form of a CFG.\footnote{\rebuttal{The CFG specification only needs to be performed manually once when the dataset is onboarded, and it may be possible to leverage LLMs to summarize the structure and generate the CFG automatically \cite{torkamani2024kajal}.}} %
For each dataset (real or synthetic), Struct-Bench then uses the user-provided CFG to construct a parse tree for each sample.

\rebuttal{\textit{Remark:} An alternate design choice could be to specify each dataset under the formalism of \emph{context-sensitive grammars (CSGs)}, which are more expressive than CFGs and can capture semantic dependencies. Specifying a CSG, however, requires significantly more domain knowledge and detail, making it a more burdensome—and potentially error-prone—process than specifying a CFG.
Hence, instead of encoding semantic dependencies as hard constraints via CSG, we empirically assess them by introducing a metric Key Node Dependency (KND) in \cref{sec:metric}. KND is simple to implement, requires less domain expertise than a full CSG specification, and statistically captures correlations in the data.}

Once the CFG is defined, the user also specifies  a set of \textit{key nodes}. {For instance, if the CFG defines a set of node types $\calS=\{\text{Query}, \text{Response}, \text{Follow-Up}\}$, the set of key nodes $\calK$ can be any subset of $\calS$.} Key nodes are expected to exhibit strong dependencies in the data. For example, in ShareGPT, we would define each query-response as a pair of key nodes.  
\rebuttal{We will assess whether the correlations among these key nodes are preserved in the synthetic data, as detailed later in \cref{sec:metric}.}
{A user can optionally not specify key nodes (in which case all nodes are treated as key nodes).

\paragraph{Privacy Constraint} 
{In this work, our goal is to evaluate differentially private synthetic data generation algorithms. 
A data generator $\dpmechanismnotation{}$ is $\bra{\epsilon, \delta}$-differentially-private if for any neighboring  datasets   $\mathcal{D}_0$ and $\mathcal{D}_1$ (i.e., $\mathcal{D}_0$ and $\mathcal{D}_1$ differ one sample), and any set $S\subseteq \functionrange{\dpmechanismnotation{}}$, we have%
    \begin{align*}
        \probof{\dpmechanismof{\mathcal{D}_0} \in S} \leq e^\epsilon\cdot \probof{\dpmechanismof{\mathcal{D}_1} \in S}+\delta~~.
    \end{align*}
In other words, the output synthetic data distribution should not depend too much on any single sample in the input dataset. 
Today, there exist many algorithms for generating DP synthetic data (some of which can accommodate text and/or structured data) \cite{yu2021differentially,xie2024differentially,hou2024pre,tan2025synthesizing,tao2021benchmarking}. %
Struct-Bench provides a systematic way of comparing DP synthetic data generators. }

\rebuttal{Struct-Bench can also be applied to other forms of privacy-preserving synthetic data, such as those generated under frameworks like quantitative information flow \cite{smith2009foundations}, statistical maximal leakage \cite{lin2024summary, wang2024statistic}, and distribution privacy \cite{suri2021formalizing}, among others. To ensure a fair comparison, all synthetic data generation baselines should be compared under the same privacy framework.}

{
\textit{Remark:}
Private DP data synthesis and non-private data synthesis are two problems with different applications and objectives. In the private setting, the goal is typically to match synthetic data to a private dataset as closely as possible under a DP constraint \cite{tao2021benchmarking,rosenblatt2020differentially,pe,hou2024pre,xie2024differentially}.
On the other hand, if there is no privacy constraint and if one wants to match the real data as closely as possible, one should just use the real data. 
Indeed, in the non-private setting, synthetic data is typically designed to deviate from the ``real'' dataset (e.g., conditional generation of a specific class of data) \cite{hattatouglu2024synthfair,longo2025synthetic,das2022conditional,douzas2018effective,engelmann2020conditional,lin2020using}. 
In our design and evaluation, we focus on DP synthetic data, and therefore \bench{} is designed to measure the similarity between the synthetic dataset and a real dataset. However, the metrics in \bench{} could be helpful for benchmarking non-private synthetic data as well, and we leave it to future work.
}

\vspace{-1mm}
\paragraph{Summary of Inputs} In summary, the Struct-Bench framework takes as input: (1) a real dataset $\base$, \footnote{\rebuttal{The input data should be in string format. For structured data with categorical or numerical values, we convert it to a JSON object where the attribute of each column is the key, and the actual value/text is its value.}} (2) a synthetic dataset $\synthetic$, (3) a CFG that represents the  structural characteristics of the data, (4) a set of key nodes (optional) from the CFG, which represent the types of nodes whose correlations are important.
Given these inputs, Struct-Bench automatically calculates the following suite of metrics.

\subsection{Struct-Bench Metrics}
\label{sec:metric}

We report three %
types of metrics: structural, non-structural, and downstream task accuracy.

\paragraph{Structural Metrics} These are metrics that depend on the CFG in some way. \rebuttal{Within a sample, structure can be defined at the level of the whole sample (\emph{CFG Pass Rate}, \emph{Attribute Match}), groups of nodes (\emph{Key Node Dependency}), and individual nodes (\emph{Attribute Match}).}
\textit{(1) CFG Pass Rate (\cfg)}: %
This measures the fraction of samples in the synthetic dataset $\synthetic$ that parse correctly under the CFG. \textit{(2) Key Node Dependency (\kns)}:
This metric measures the semantic dependencies between ``key node pairs'', {which are pairs of nodes believed to have a meaningful relation;\footnote{\rebuttal{We measure dependencies between node pairs and do not consider higher-order relationships, as they are more computationally intensive and the dependency functions can vary across different scenarios.}} for example, in a question-and-answer dataset, we would expect that each associated question and answer pair should have a strong correlation}. Programmatically, users specify pairs of key nodes with \texttt{Tregex} \cite{levy2006tregex}, a tool for matching regular expressions on trees. 
Typically, we can measure the dependencies by cosine similarity of the node embeddings, while one can also adopt LLM as a judge or other dependency functions defined by the users. For a dataset, we can construct a distribution of dependencies of node pairs in the same pattern. To evaluate the similarity of the node dependencies captured by the private and synthetic dataset, we then calculate the distributional distance, e.g., Wasserstein-2 distance, between the private and synthetic dependency distributions. \textit{(3) Attribute Match (\knd)}: %
This metric measures the distributional distance of sample-level attributes (e.g., number of nodes) or node-level attributes (e.g., node token length) between the private and synthetic datasets. The attribute can be a statistical property or a semantic property, {and it can be derived either from an explicit attribute function or through human annotation or LLM-based labeling.} The distributional distance can be Wasserstein-2 distance if the attribute is numeric  or total variation distance if the attribute is  categorical. %
\rebuttal{The precise definitions and instantiation guidelines for \kns{} and \knd{} are provided in \cref{app:metric}.}
For our structural metrics, higher \cfg{} indicates better performance, while \kns{} and \knd{} are the opposite.%

\paragraph{Non-Structural Metrics}
\label{sec:non-structral_metric} Following prior precedent \cite{tao2021benchmarking,arnold2020really}, non-structural metrics are per-sample metrics that do not rely on the CFG, i.e. they are unrelated to the structure of the data.
These metrics quantify the similarity between the content of the generated synthetic data and the real/private data. %
Moreoever, as in prior works on unstructured DP synthetic data \cite{kynkaanniemi2019improved,xie2024differentially}, %
we report the precision (\emph{\precision{}}) and recall (\emph{\recall{}}) for each synthetic dataset. Roughly, \precision{} (resp. \recall{}) calculates the proportion of synthetic (resp. private) samples whose embedding distance to a private (resp. synthetic) sample is smaller than this sample's $k$-th nearest neighbor within its own dataset. \precision{} evaluates the average semantic quality of the synthetic samples, and \recall{} assesses the semantic diversity of the samples \cite{kynkaanniemi2019improved}.

\paragraph{Downstream Evaluations (DE) Based on Label Prediction} 
The eventual goal of synthetic data is typically a downstream task, e.g., training a machine learning (ML) model. 
Struct-Bench allows  users to design their own downstream label prediction tasks. Our  pipeline  includes label generation, downstream model training, and evaluation. \emph{(1) Label Generation}: The evaluation pipeline requires labels both for the synthetic and real data. Labels can either be generated by a human or we can use large language models (LLMs) to simulate a human labeler. In our evaluation, we adopted GPT-4o to label the samples, but the Struct-Bench codebase gives users flexibility to choose a different LLM. \rebuttal{We provide guidelines for the label generation process in \cref{app:label_generation}.} \emph{(2) Downstream Model Training}: To conduct label prediction, we fine-tune a language model based on the synthetic dataset and its label. Since the samples may have long text, we adopt Longformer \cite{longformer} in this paper. \emph{(3) Evaluation:} We evaluate the downstream task by calculating the prediction accuracy (\downstream{}) on a held-out test set from the real data. 
{This is commonly done in synthetic data evaluations, and is known as the train-synthetic-test-real (TSTR) framework \cite{van2023synthetic,qian2024synthetic,lin2020using}}.\footnote{{Some evaluation frameworks compare TSTR with train-real-test-real (TRTR) for a self-contained evaluation \cite{yuan2024multi}. However, since we are comparing different synthetic data generation algorithms \emph{against each other}, we compute only TSTR in this case, which is slightly more interpretable.}}

\section{%
Benchmarking  Differentially Private Synthetic Data %
}
\label{others} %
To demonstrate the utility of the \name framework, we instantiate and implement it on a set of seven datasets and four DP synthetic data generation methods.

\subsection{Struct-Bench Datasets}
\label{headings}

\begin{table}[t] %
  \caption{List of datasets with descriptions, key nodes, and sizes of real and generated data.%
  }
  \label{tab:new_dataset_list_expanded}
  \centering
  \begin{tabular}{lllcc}
    \toprule
            &                     &                       & \multicolumn{2}{c}{Number of Samples} \\
    \cmidrule(lr){4-5}
    Dataset & Description         & Key Nodes             & Real & Generated \\
    \midrule
    ShareGPT         & Human–GPT conversations            & query, response & 3\,000 & 600 \\
    ICLR             & ICLR paper reviews \& rebuttals     & review, rebuttal, comment & 3\,000 & 300 \\
    Adult            & Census dataset                      & native country, workclass & 50\,000 & 31\,561 \\
    Water            & Water-bottle reviews                & title, review & 25\,000 & 20\,000 \\
    Arena            & Chatbot-Arena conversations         & conversation 1 \& 2 & 25\,000 & 20\,000 \\
    Reviews & Synthetic product reviews                                   & review, rating & 2\,000 & 2\,000 \\
    Grounding & Synthetic grounded QA                                   & query, response & 2\,500 & 2\,000 \\
    \bottomrule
  \end{tabular}
  \vspace{-2mm}
\end{table}

We include three types of datasets in \bench{} (details in \cref{app:dataset_description}).
\vspace{0mm}
\paragraph{Real-World Datasets with Graph-Structured Dependencies}
{We use \textbf{ShareGPT} \cite{anon_sharegpt_vicuna_unfiltered_commit_bcd32a7} and \textbf{ICLR} 2024 paper reviews \cite{ICLR2024}, which represent two typical real-world examples of graph-structured datasets with significant differences: (1) Content: ShareGPT consists of multi-round conversations between a user and AI agents (GPT), whereas the ICLR 2024 paper review dataset contains reviews, rebuttals, and comments of the papers submitted to ICLR 2024 between the authors and multiple reviewers; (2) Topic: ShareGPT conversations cover a wide range of open-domain topics, while the ICLR 2024 paper review dataset is specific to AI research areas; (3) Structure: ShareGPT conversations follow a linear structure -- each user query is followed by a GPT response, continuing sequentially; while ICLR posts exhibit a tree structure, in which a paper receives multiple reviews, followed by corresponding rebuttals and subsequent discussions between authors and reviewers. Both datasets have diverse semantics and logic and clear data structures with at least two types of nodes. 
Notably, the ICLR 2024 paper review dataset was released after the training data cut-off date for the LLMs we evaluate.}

\vspace{-1.5mm}
\paragraph{Real-World Tabular Datasets} \rebuttal{Although there exist benchmarks for tabular data \cite{arnold2020really,tao2021benchmarking}, they do not naturally extend to natural language fields. We evaluate \name on three tabular datasets. 
Two of them (Water and Arena) contain textual fields while the third dataset (Adult) contains only numerical and categorical values. We include the Adult dataset to demonstrate that Struct-Bench can be applied to non-textual data as well, but note that this data type is not a main focus of our work.} %
The \textbf{Water} \cite{Tharunmss_WaterBottle_Flipkart_Kaggle} dataset contains reviews of water bottles, the \textbf{Arena} \cite{zheng2023judging} dataset contains pairs of human-model conversations, and the \textbf{Adult} \cite{BeckerKohavi1996} dataset contains census data.

\vspace{-1.5mm}
\paragraph{Synthetic Datasets with Controllable Data Attributes} %
{To control the complexity of the data structure and semantics more explicitly,} we construct two synthetic datasets named \textbf{Synthetic Reviews} and the \textbf{Synthetic Grounding Dataset}. The Synthetic Reviews dataset contains reviews with varying sentiments and review scores, and the Synthetic Grounding dataset contains source documents and a question-answering based on them. %

We show the key nodes of each dataset and the sizes of the real and generated data in \cref{tab:new_dataset_list_expanded}, and defer the detailed data modeling to \cref{app:data_modeling}.

\subsection{\bench{} Synthetic Data Generation Baselines}
\label{sec:baselines}

Since we focus on structured datasets that contain natural language, we select LLM-based DP synthetic data generators as our baselines, specifically, Private Evolution (PE) \cite{lin2023differentially,xie2024differentially,lin2025differentially}, DP model fine-tuning \cite{yu2021differentially,wutschitz2022dp,yue2022synthetic}, and some variants. While several methods have been proposed to generate DP tabular data using LLMs \cite{afonja2024dp,tran2024differentially}, they are typically limited to handling only numerical or categorical values and cannot generate structured data that incorporates natural language.  %

\vspace{-1.5mm}
\paragraph{Private Evolution (\PE) \cite{lin2023differentially,xie2024differentially,lin2025differentially}} PE is a leading training-free DP synthetic data generation algorithm that makes use of foundation models pre-trained on public data \cite{lin2023differentially,xie2024differentially,lin2025differentially,hou2024pre,gong2025dpimagebench,wang2025synthesize}. %
PE first uses a Random API to generate initial samples from the foundation language model. Then, it iteratively: (1) constructs a differentially private (noisy) voting histogram based on private samples voting for their nearest synthetic counterparts; (2) draws samples according to this histogram; and (3) creates new samples with a Variation API that generates perturbed versions of the original samples. 
{We use the PE variant, Augmented Private Evolution (Aug-PE) \cite{xie2024differentially}, provided by the Private Evolution library;\footnote{\url{https://github.com/microsoft/DPSDA}} our only modification is to choose the prompt for the Random and Variation APIs.}
\vspace{-1.5mm}
\paragraph{Instruction Following (\IF)} %
    \IF{} uses foundation language models to generate samples based on a prompt that outlines the required data structure, without incorporating any signals or data from the private dataset, and therefore $\epsilon=0$.
    This can be viewed as a zero-shot version of PE, i.e., it is equivalent to using only the Random API in \PE{} to generate samples.
\vspace{-1.5mm}
\paragraph{DP Fine-Tuning (\DPFT) \cite{yu2021differentially,wutschitz2022dp,yue2022synthetic}} \DPFT{} adopts DP stochastic gradient descent (DP-SGD) \cite{abadi2016deep} to fine-tune the language model on  the next token prediction task. We generate synthetic data unconditionally from the fine-tuned model. This simple baseline remains competitive when training is allowed \cite{gong2025dpimagebench}.  \rebuttal{Since \PE{} generates synthetic data based on the instructions in a prompt in the Random and Variation APIs, we also include a variant of \DPFT{} that conditionally generates samples according to {the same} instructions after DP fine-tuning, which we refer to as \emph{Instruct DP-FT}. {We provide details of this instruction fine-tuning in \cref{app:results}.}}
\vspace{-1.5mm}
\paragraph{Real Data Fine-Tuning (\FT)} \FT{} directly fine-tunes the language model without any privacy guarantees, and therefore $\epsilon=\infty$. This is a best-case scenario for %
\DPFT{}. \rebuttal{We also include \emph{Instruct \FT{}}, which utilizes instruction-guided conditional generation. %
} %

\medskip
\rebuttal{Since \DPFT{} and \FT{} require model training, we are limited to open-source models and thus use only GPT-2. In contrast, PE and IF do not require access to model weights, making them compatible with models available exclusively through APIs; therefore, we evaluate them on both GPT-2 and the state-of-the-art GPT-4o.}
\rebuttal{We further evaluate our baselines on additional foundation models in \cref{app:baseline_llama2}.%
} %
\PE{} is run for  10 iterations, and \DPFT{} and \FT{} are run for 20 epochs.  We vary the privacy budget $\epsilon$ as $1, 2, 4$ for \PE{} and \DPFT{}, and set $\delta=0$.

\subsection{Experimental Results}

We present the results of benchmarking the DP synthetic data generation methods under ShareGPT and ICLR datasets with $\epsilon=4$ in \cref{tab:senteval_baselines}, and defer the results on other datasets to \cref{app:results_table}. %
We illustrate the performance of the baselines on \cfg{} and \recall{} across all datasets in radar plots shown in \cref{fig:main_radar}. %
We specify the metrics reported and the whole set of evaluation metrics we adopt in \cref{app:metrics}; {additional results for $\epsilon=2$ and $\epsilon=1$ the on the ShareGPT and ICLR datasets can be found in \cref{app:baseline_compare}.%
} 
\cref{tab:senteval_baselines,fig:main_radar} highlight several main takeaways: 
\begin{itemize}[leftmargin=*]
\item {\textit{No single metric fully describes synthetic data quality.} For a single algorithm and dataset, some metrics can be high, while others remain low (e.g., see \cfg{} and \recall{} in \cref{fig:main_radar}). This further motivates the need for \name, which aggregates many diverse metrics. }
\item{\textit{Existing DP synthetic data generators struggle to learn complicated data structures.} All baselines achieve a \cfg{} score below $0.2$ on the ICLR dataset, which features more node types and a significantly more intricate graph structure than ShareGPT.}
\item{\textit{DP fine-tuning alone cannot learn structure.} At $\epsilon=4$,  it achieves a \cfg{} of 0 on all of our datasets.  {Even at $\epsilon = \infty$, it fails to learn structural information on all datasets except ShareGPT, where it achieves a \cfg{} of 0.53; this is likely because ShareGPT contains fewer formatting tokens compared to other datasets (e.g., JSON tags in tabular datasets).} %
}%
\item{\textit{\PE{} and \IF{} learn structure at the expense of semantic performance.} Although \PE{} and \IF{} reliably capture the data structure \rebuttal{with SOTA models}, they suffer from poor semantic performance (low \recall{}).}
\item{\rebuttal{\textit{The performance gap between \PE{} and \DPFT{} {may} arise from the foundation models they employ and the use of instruction-guided generation.} Unlike \DPFT{}, \PE{} does not require model training, which allows us to leverage SOTA models in the Random and Variation APIs. In contrast, \DPFT{} relies on fine-tuning, and we thus need to use smaller, open-source models due to computational restrictions and the weight access requirement. Nevertheless, with instruction-guided conditional generation, Instruct \DPFT{} achieves performance comparable to \PE{} across most metrics on both ShareGPT and ICLR, {when both use the same foundation model (GPT-2)}.}}
\end{itemize}

\begin{table}[htbp]
\vspace{-1.5mm}
  \caption{DP synthetic data generation benchmarking results on Struct-Bench with $\epsilon=4$. \rebuttal{All baselines use GPT-2 unless otherwise specified.}} %
  \label{tab:senteval_baselines}
  \centering
  \begin{tabular}{llcccccc}
    \toprule
            &           & \multicolumn{3}{c}{\hspace{-1mm}Structural Metrics} & \multicolumn{2}{c}{Non-Structural Metrics} & \multicolumn{1}{c}{\hspace{-2mm}DE} \\
    \cmidrule(r){3-5}\cmidrule(r){6-7}\cmidrule(r){8-8}
    Dataset & Baseline  & \hspace{-3mm}\cfg{} $\uparrow$ & \kns{} $\downarrow$ & \knd{} $\downarrow$ & \hspace{-2mm}\precision{} $\uparrow$ & \hspace{-2mm}\recall{} $\uparrow$ & \hspace{-2mm}\downstream $\uparrow$ \\
    \midrule
ShareGPT        
& IF ($\epsilon=0$) & 0.03 & 0.07 & 41.86 & 0.64 & 0.31 & \hspace{-2mm}0.28 \\
& \makecell[l]{\IF{} ($\epsilon=0$)\\\ (\textit{GPT-4o})}          & \textbf{0.87} & 0.06 & 43.85 & 0.72 & 0.26 & \hspace{-2mm}0.38 \\
    & \FT{} ($\epsilon=\infty$)  %
    & 0.53 & \textbf{0.03} & 52.70 & 0.76 & \textbf{0.66} & \hspace{-2mm}0.37 \\
    & \makecell[l]{Instruct FT\\ \ ($\epsilon=\infty$)} & 0.59 & 0.04 & \textbf{30.70} & 0.80 & 0.54 & \hspace{-2mm}0.37 \\
    \cmidrule(r){2-8}
    & \DPFT{}       & 0 & - & - & 0.02 & 0 & \hspace{-2mm}- \\
    & Instruct DP-FT & 0.55 & 0.18 & 32.59 & 0.77 & 0.31 & \hspace{-2mm}0.35 \\
    & PE & 0.57 & 0.12 & {34.51} & 0.78 & 0.33 & \hspace{-2mm}0.38 \\
    & \PE{}  (\textit{GPT-4o})       & 0.86 & 0.07 & 38.17 & \textbf{0.81} & 0.15 & \hspace{-2mm}\textbf{0.39} \\
    \midrule
ICLR    
& IF ($\epsilon=0$) & 0.09 & \textbf{0.11} & 207.62 & 0.66 & 0.28 & \hspace{-2mm}0.39 \\
& \makecell[l]{\IF{} ($\epsilon=0$)\\\ (\textit{GPT-4o})}          & 0.17 & 0.26 & \textbf{204.80} & 0.84 & 0.03 & \hspace{-2mm}0.47 \\
    & \FT{} ($\epsilon=\infty$)  %
    & 0 & - & - & 0.71 & \textbf{0.47} & \hspace{-2mm}0.46 \\
    & \makecell[l]{Instruct FT\\ \ ($\epsilon=\infty$)} & 0.09 & 0.16 &	208.37 &	0.77 &	0.29 & \hspace{-2mm}0.51 \\
    \cmidrule(r){2-8}
    & \DPFT{}        & 0 & - & - & 0 & 0 & \hspace{-2mm}0.18 \\
    & Instruct DP-FT & 0.08 &	0.22 &	237.52 &	0.49 &	0.18 & \hspace{-2mm}0.40 \\
    & PE & 0.10 & 0.20 & 248.73 & 0.49 & 0.20 & \hspace{-2mm}0.41 \\
    & \PE{} (\textit{GPT-4o})        & \textbf{0.19} & 0.26 & 240.94 & \textbf{0.98} & 0.02 & \hspace{-2mm}\textbf{0.52} \\
    \bottomrule
  \end{tabular}
  \vspace{-1mm}
\end{table}

\begin{figure}[htbp]
\vspace{-1mm}
    \centering
    \begin{subfigure}{0.45\textwidth}
         \centering
    \includegraphics[width=1\linewidth]{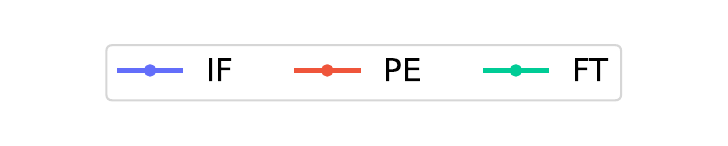}
\end{subfigure}
\vspace{-5mm}

\begin{subfigure}{0.46\textwidth}
         \centering
    \includegraphics[width=1\linewidth]{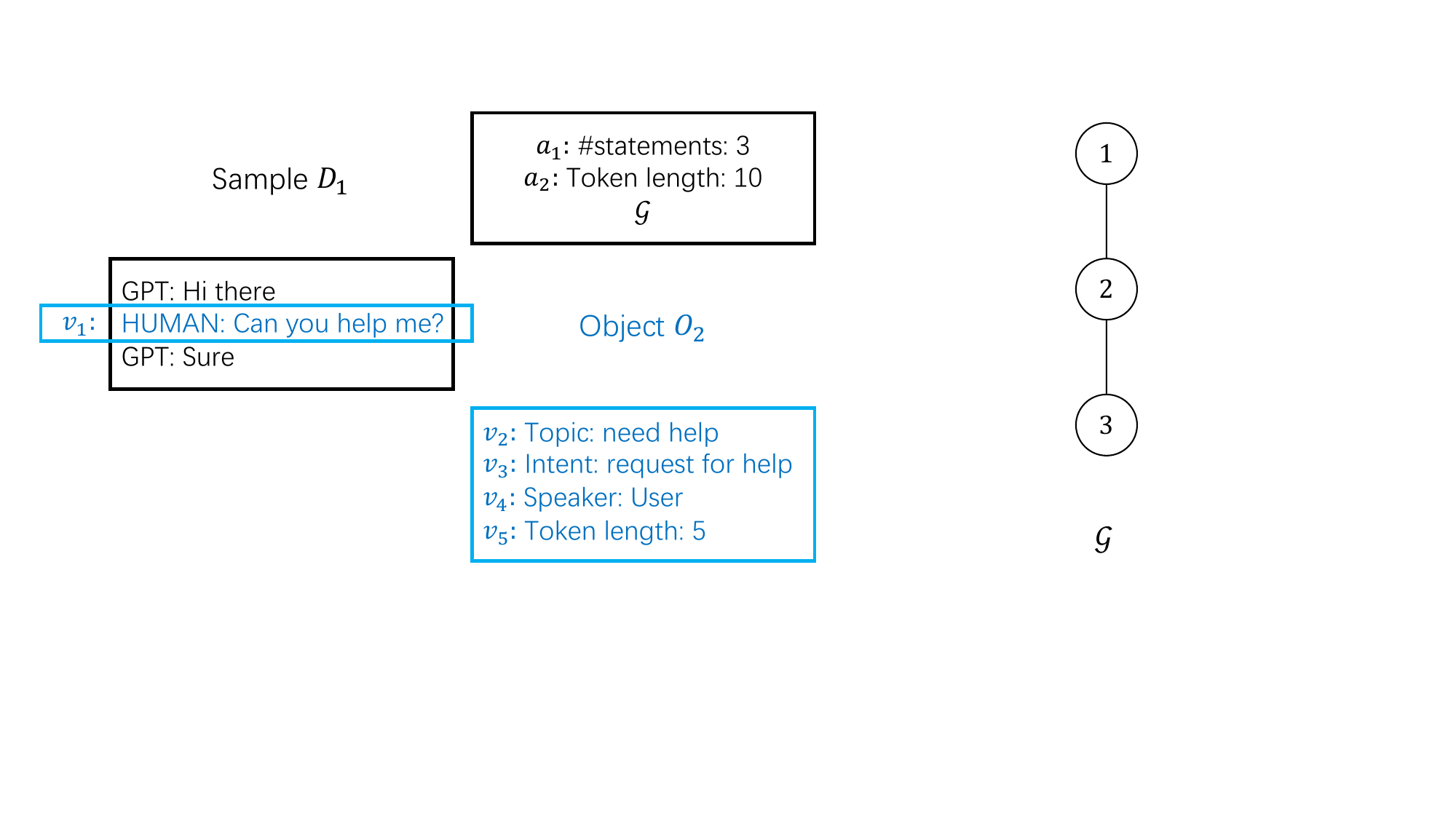}
    \caption{\cfg{} $\uparrow$}
\end{subfigure}
\begin{subfigure}{0.46\textwidth}
         \centering
    \includegraphics[width=1\linewidth]{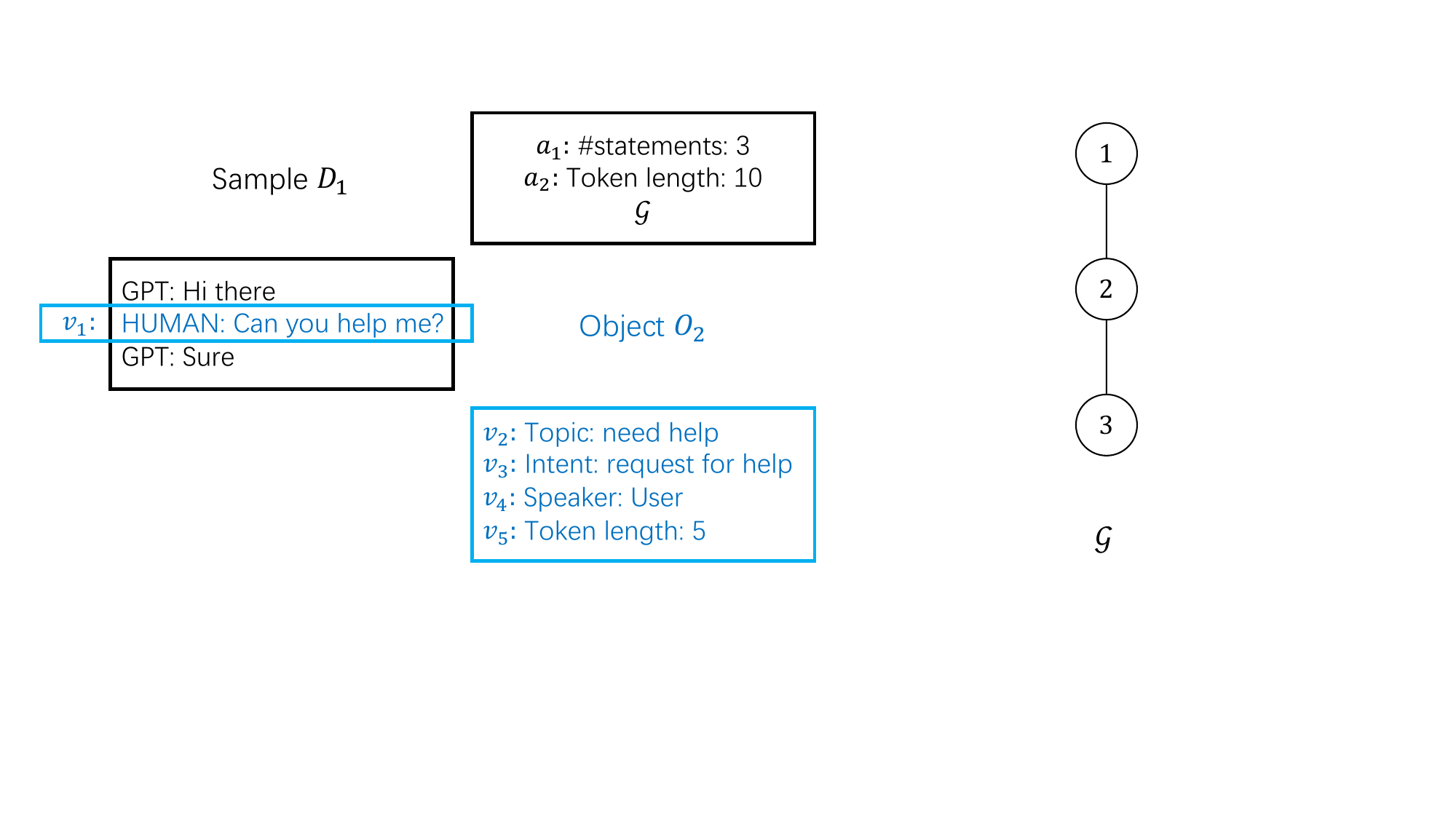}
    \caption{\recall{} $\uparrow$}
\end{subfigure}
    \caption{\cfg{} and \recall{} of baselines on \bench{} with different datasets. %
    {While frontier models can capture the syntactic structure of many of our datasets (i.e., \cfg{} is high), existing DP synthetic data generation techniques do not capture the semantic diversities %
    (KNN-Recall).}  \rebuttal{With GPT-4o,} both \IF{} and \PE{} achieve perfect \cfg{} on tabular and synthetic datasets, while \FT{} \rebuttal{with GPT-2} fails to learn the correct data format. However, \recall{} of all baselines are near $0$, indicating poor semantic performance on tabular and synthetic datasets.
    {We exclude \DPFT{} as it attains a score of zero on both metrics for all datasets. \rebuttal{%
    Note that Instruct \DPFT{} and Instruct \FT{} may achieve stronger results than the \DPFT{} and \FT{} variants shown here, as reported in \cref{tab:senteval_baselines}.}  %
    }
    }
    \label{fig:main_radar}
    \vspace{-3mm}
\end{figure}

\section{Case Studies}
\label{sec:main_case_study}

In this section, we demonstrate how users can leverage \bench{} to better understand and improve \PE{}. %
\rebuttal{We focus on \PE{} because, unlike \DPFT{}, it does not require \emph{training} on private data, offering both efficiency and qualitative advantages \cite{xie2024differentially,hou2024pre}. We use the the ShareGPT dataset for this case study as it is a real-world dataset that is semantically diverse and has a larger token length for each node relative to our tabular or synthetic datasets. While GPT-4o might achieve stronger performance, we adopt Llama3-8b as the foundation model in this section since it is more cost-efficient for our experiments and, importantly, more affordable and accessible to end users.}

\begin{wrapfigure}{r}{0.38\textwidth}
  \centering
  \vspace{-1mm}
  \includegraphics[width=\linewidth]{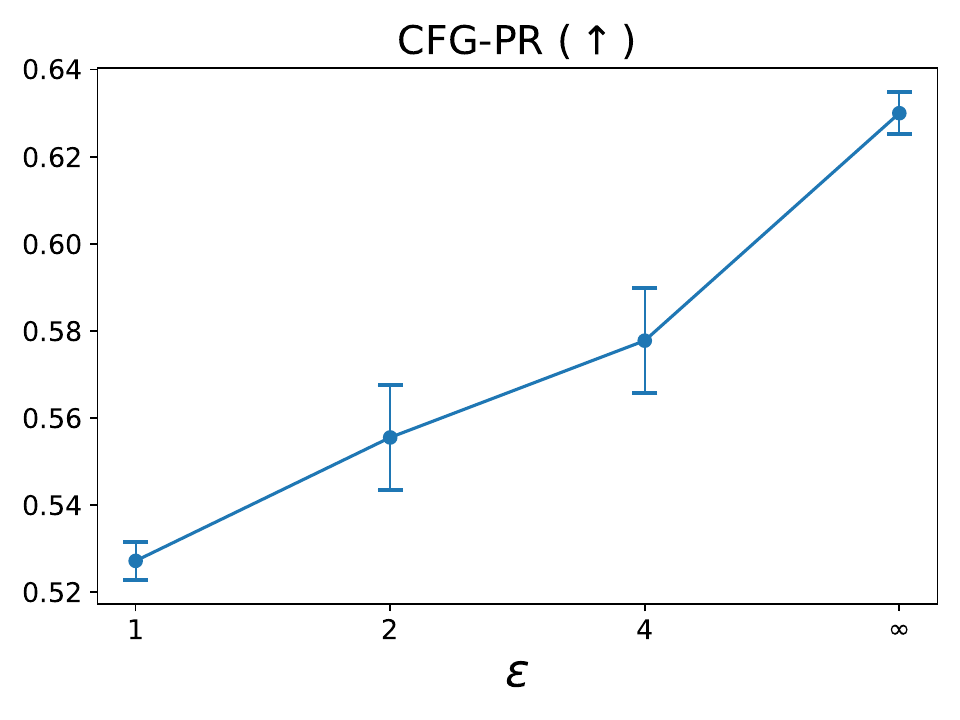}
  \caption{%
  \cfg{} of vanilla \PE{} {on ShareGPT}. Each data point is averaged over three independent trials. %
  \cfg{} is low for all $\epsilon$. %
  }%
  \label{fig:case_study_cfg}
  \vspace{-1mm}
\end{wrapfigure}
We mainly focus on improving structural validity and semantic diversity of the \PE{} synthetic data in this section, and defer a more thorough analysis as well as methods on improving node dependency (\kns{}) to \cref{app:case_study}.

\paragraph{Problem 1: Structural Validity (\cfg{}) is low.}

Structural validity, i.e., \cfg{}, {is a critical metric, as many downstream applications on structured datasets 
expect data to be formatted in a particular way for compatibility with utilities and dataset-specific pipelines.} %
\cref{fig:case_study_cfg} shows that the \cfg{} of vanilla \PE{} is below 60\% when $\epsilon\leq 4$, and only achieves  $\sim63\%$ in the non-private setting (i.e., we run PE with no added noise).
This suggests that with smaller foundation models (e.g., Llama3-8B), vanilla PE inherently cannot capture even simple structural constraints.

\noindent \textbf{Solution 1: LLM-Assisted reformatting can improve CFG compliance.}
To improve structural validity, we introduce a \emph{reformatting} feature to the Random and Variation APIs by prompting LLMs to explicitly check and reformat CFG-invalid 
samples (see \cref{app:reformat_prompt}). %
For example, if PE generates `HUMAN: How are 
you?', the model detects the missing response and reformats it to `HUMAN: How are you? GPT: I'm fine.'

\begin{wrapfigure}{l}{0.4\textwidth}
  \centering
    \vspace{0mm}
    \includegraphics[width=1\linewidth]{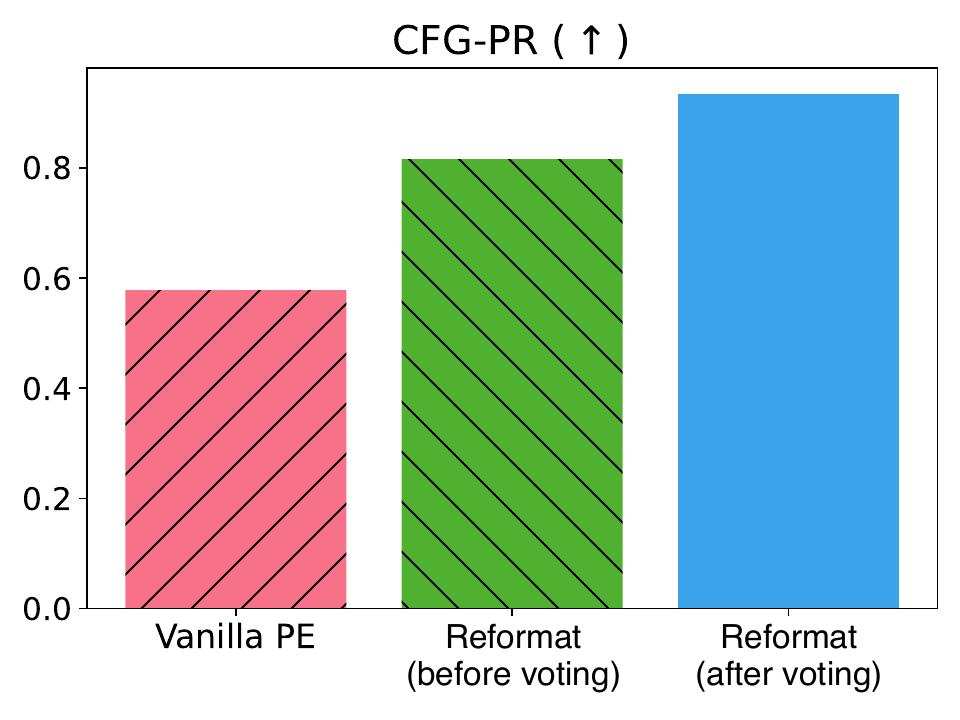}
    \caption{\cfg{} of vanilla \PE{} and \PE{} with CFG reformat under $\epsilon=4$.%
    }
    \label{fig:case_study_reformat_cfg}
    \vspace{0mm}
\end{wrapfigure}
As described in \cref{sec:baselines}, \PE{} generates a set of samples and then select those with high qualities by private voting in each iteration. %
Sample reformatting can happen before or after the 
\PE{} private voting process. 
We compare both reformatting
methods with vanilla \PE{} %
at $\epsilon=4$ in \cref{fig:case_study_reformat_cfg}. %
We see that the \cfg{} of both our methods increases by over 20\% compared with vanilla \PE{}, and reformatting after {the \PE{} private voting procedure does the best.} 

Reformatting %
can help enforce structural correctness but may come at the cost of semantic integrity. For example, as illustrated in \cref{fig:reformat}, if \PE{} generates “How are you? I'm fine. Thanks.”, the model may detect missing format tokens and reformat it to “HUMAN: How are you? I'm fine. GPT: Thanks!” While the reformatted version follows a valid structure, its semantics are flawed—the user’s query includes part of the response. As a result, in the reformatting-before-voting setting, the voting process may become biased against structurally valid but semantically distorted samples. %
In contrast, reformatting-after-voting directly reformats only the most highly-voted samples, which are either used as a final output (at the last iteration of PE) or utilized as seeds in the next PE iteration without further selection, resulting in a higher \cfg{}.

\begin{figure}[htbp]
    \centering
    \includegraphics[width=\linewidth]{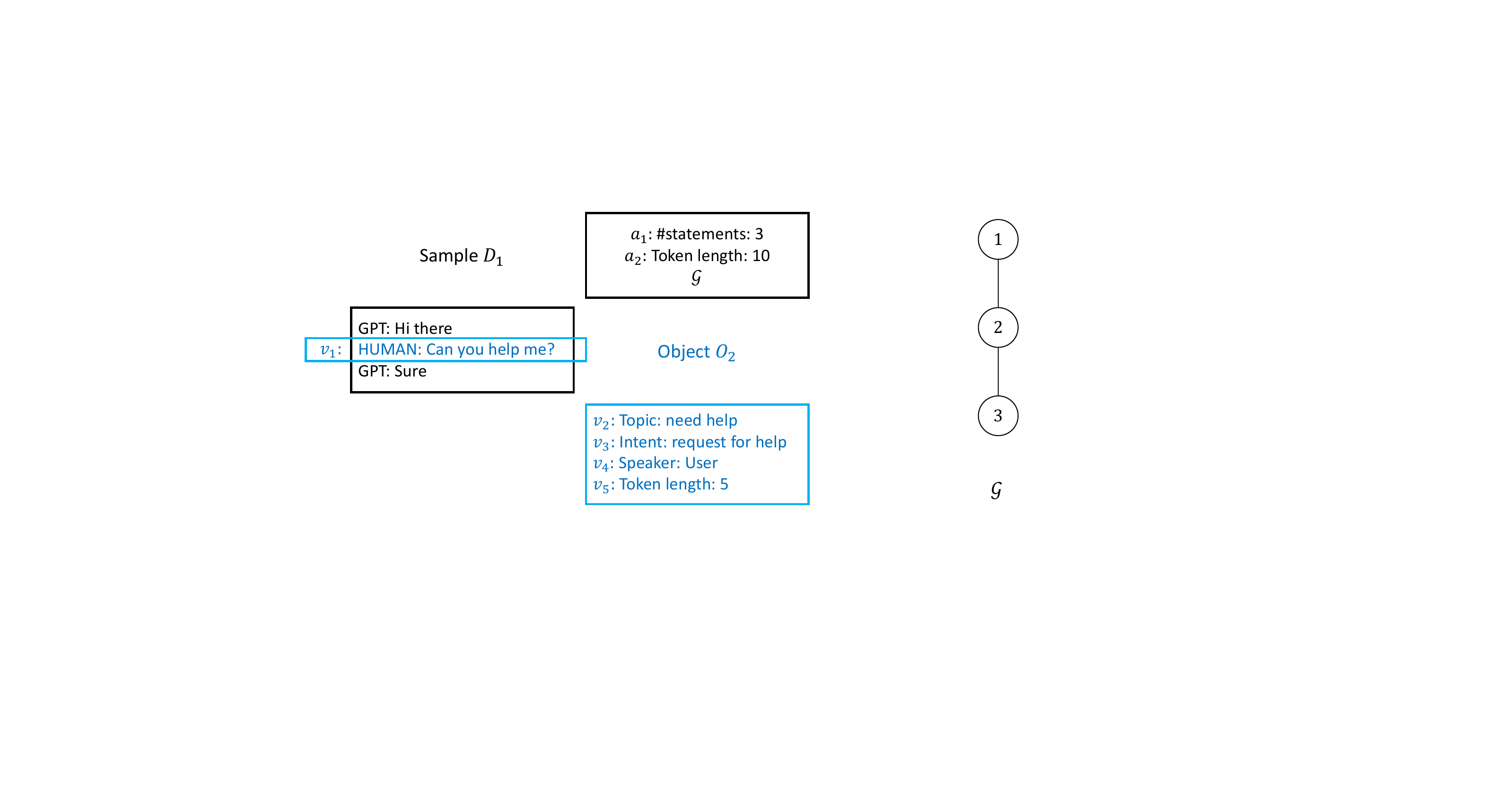}
    \caption{{Illustration of reformatting-before-voting on the ShareGPT dataset. The syntactically-correctly reformatted sample follows a valid structure but has flawed semantics—the user’s query includes part of the response. In contrast, the incorrectly reformatted sample preserves semantic integrity. In these cases, the voting process can become biased toward samples that are semantically consistent but structurally invalid.%
} 
    }
    \label{fig:reformat}
\end{figure}

\paragraph{Problem 2: Semantic diversity (\recall{}) is low.}
As described in \cref{sec:non-structral_metric}, \precision{} measures the semantic quality of the generated samples, and the \recall{} measures how well the semantic diversity of the private dataset is captured in the synthetic data. As we see in \cref{fig:case_study_precision_recall}, %
as $\epsilon$ increases, i.e., with looser privacy constraints, the \precision{} of vanilla \PE{} increases from $0.56$ to $0.69$, 
\begin{wrapfigure}{r}{0.4\textwidth}
  \centering
    \vspace{0mm}
    \includegraphics[width=\linewidth]{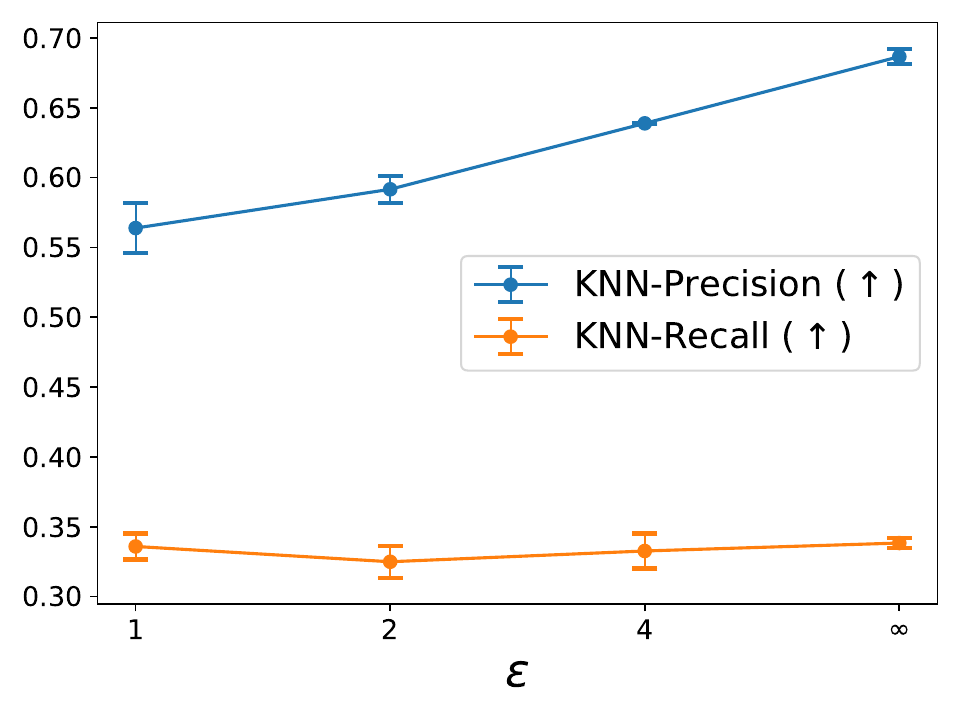}
    \caption{\precision{} and \recall{} of vanilla \PE{} {on ShareGPT} under different privacy guarantees. Both are low.}
    \label{fig:case_study_precision_recall}
    \vspace{0mm}
\end{wrapfigure}
increases from $0.56$ to $0.69$, while \recall{} remains very low, around $0.35$. This suggests that  vanilla \PE{} focuses on semantic quality while sacrificing diversity. %

\paragraph{Solution 2: Node extraction \& auto-generation can improve semantic diversity.} %
In the Variation API, PE generates new samples by %
first masking a subset of the original text and then using the LLM to fill in the blanks based on the remaining context. %
This process largely preserves the original meanings, which limits semantic diversity. %
For example, if a conversation is about weather, and its masked version retains keywords like `cloudy' or `rainy', the blank-filled new sample will likely still be about weather, rather than an unrelated topic like dogs.

To improve semantic diversity, we propose a variant of PE that \emph{extracts} specific nodes for blank-filling and then allows the language model to auto-generate the remaining nodes %
with fewer semantic constraints in the Variation API.%
{ We call this pipeline ``Node extraction and auto-generation", and illustrate an example on the ShareGPT dataset in \cref{fig:extract}.} The user must specify which nodes are to be extracted---for instance, these could be roots in the parse tree of the CFG. The remaining nodes are auto-generated by feeding the extracted nodes (with blank-filled variations generated) into an LLM and asking it to generate the remaining nodes in the sample. Due to the post-processing property of DP, this pipeline does not incur any additional privacy cost.

\begin{figure}[htbp]
    \centering
    \includegraphics[width=\linewidth]{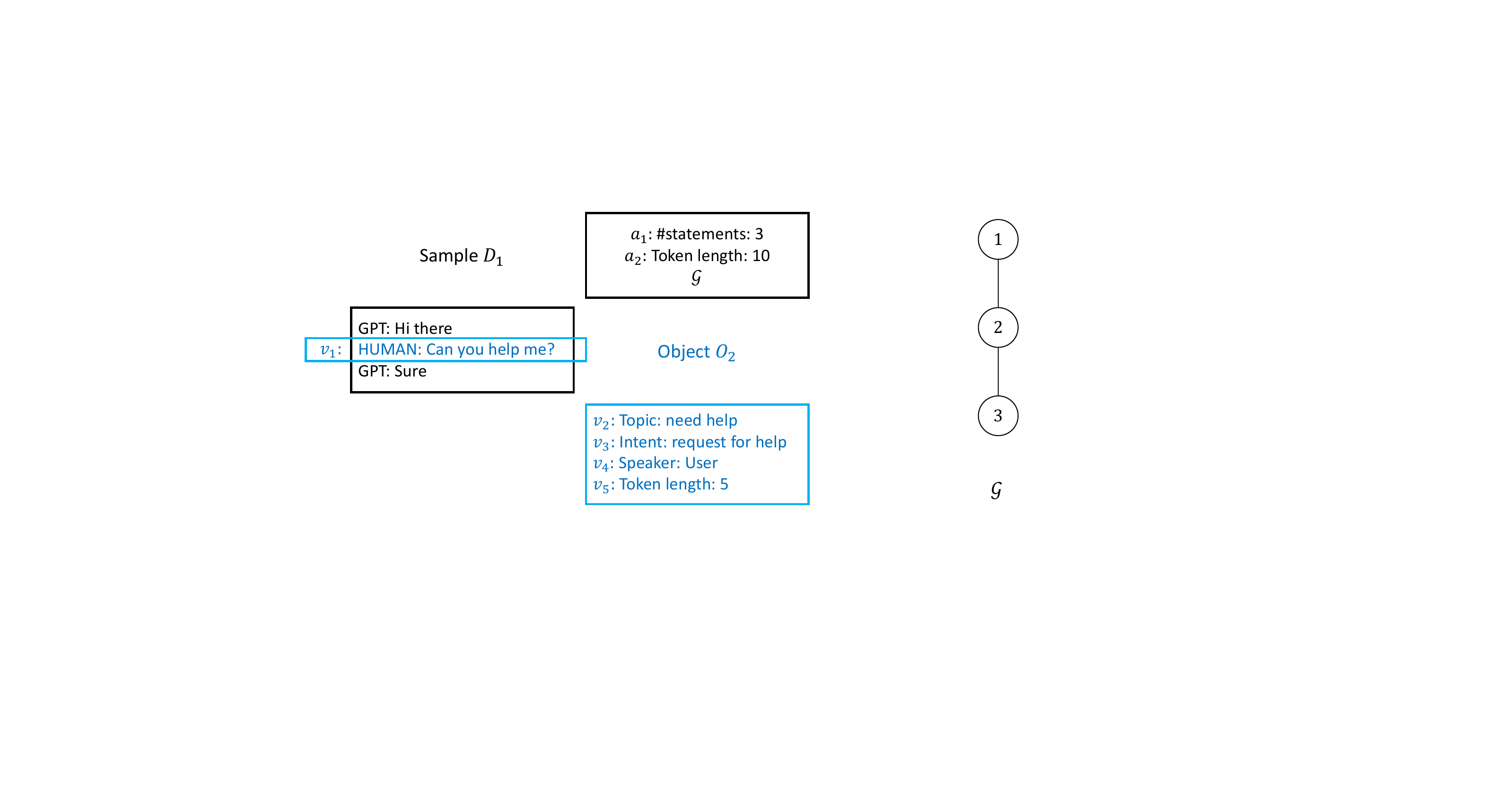}
    \caption{{Example executions of the Variation API of vanilla \PE{} and \PE{} with node extraction and auto-generation on the ShareGPT dataset. Vanilla \PE{} generates new samples by first masking a subset of the original text and then using the LLM to fill in the blanks based on the remaining context. In contrast, \PE{} with query extraction and auto-generation first extracts node(s) from the conversation; in this example, the extracted node is the ``Query".  It then conducts blank-filling only on the extracted Query node. The language model then generates responses conditioned on the query. This produces fewer semantic constraints, allowing this variant of PE to generate more semantically-diverse samples.}}
    \label{fig:extract}
\end{figure}

We compare the performance of vanilla \PE{} and \PE{} with node extraction in 
\cref{fig:case_study_extract}. 
{Recall that the ShareGPT dataset has only two types of nodes: query and response nodes.} In \cref{fig:case_study_extract}, 
we consider two variants of the node extraction method: (1) extract all queries and auto-generate all responses (listed as Extract Query), and (2) extract all responses and auto-generate all queries (shown as Extract Response). 
We show three metrics: KNN-Recall, KNN-Precision, and \cfg{}.

\cref{fig:case_study_extract} shows that Extract Query not only improves the \recall{} but also the \precision{}, (i.e., semantic quality) as auto-generation also ensures the semantic meaning is more consistent and natural across nodes. However, Extract Response does not improve \recall{} as the semantic diversity depends mainly on responses, while queries are fairly constrained for a given response. This indicates that the type of nodes extracted is crucial to the performance of semantic diversity. 
Additionally, since %
{the formatting tokens around the extracted critical nodes will not be accidentally modified by Variation API, the \cfg{} of \PE{} with node extraction is also higher than vanilla \PE{}.}%

\begin{figure}[htbp]
    \centering
\begin{subfigure}{0.55\textwidth}
         \centering
    \includegraphics[width=1\linewidth]{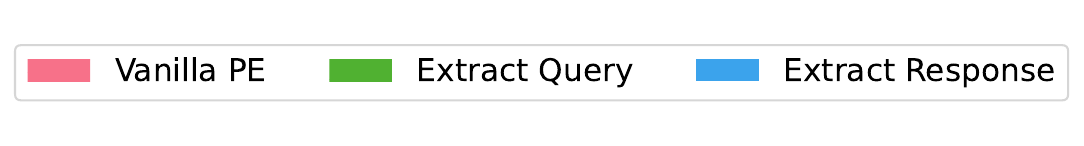}
\end{subfigure}
\vspace{-1mm}

\begin{subfigure}{0.3\textwidth}
         \centering
    \includegraphics[width=1\linewidth]{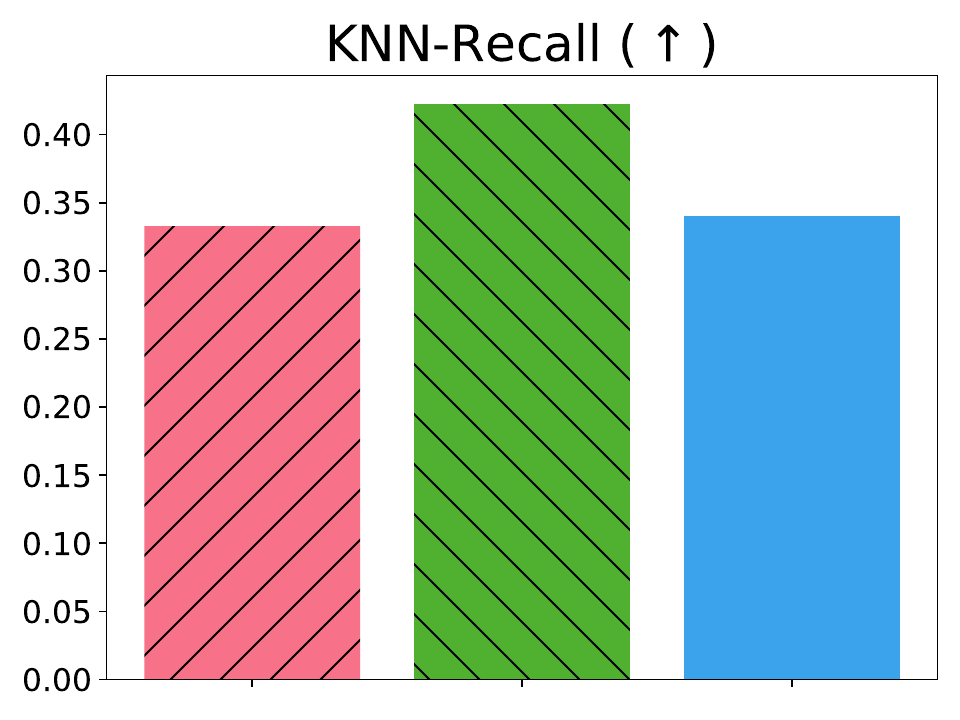}
\end{subfigure}
\begin{subfigure}{0.3\textwidth}
         \centering
    \includegraphics[width=1\linewidth]{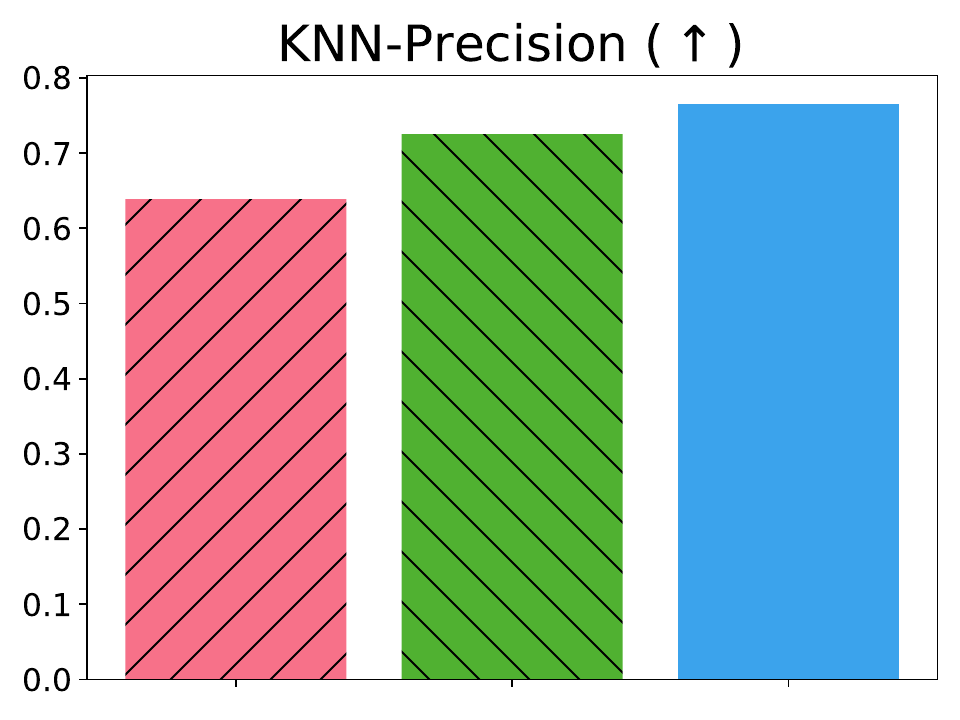}
\end{subfigure}
\begin{subfigure}{0.3\textwidth}
         \centering
    \includegraphics[width=1\linewidth]{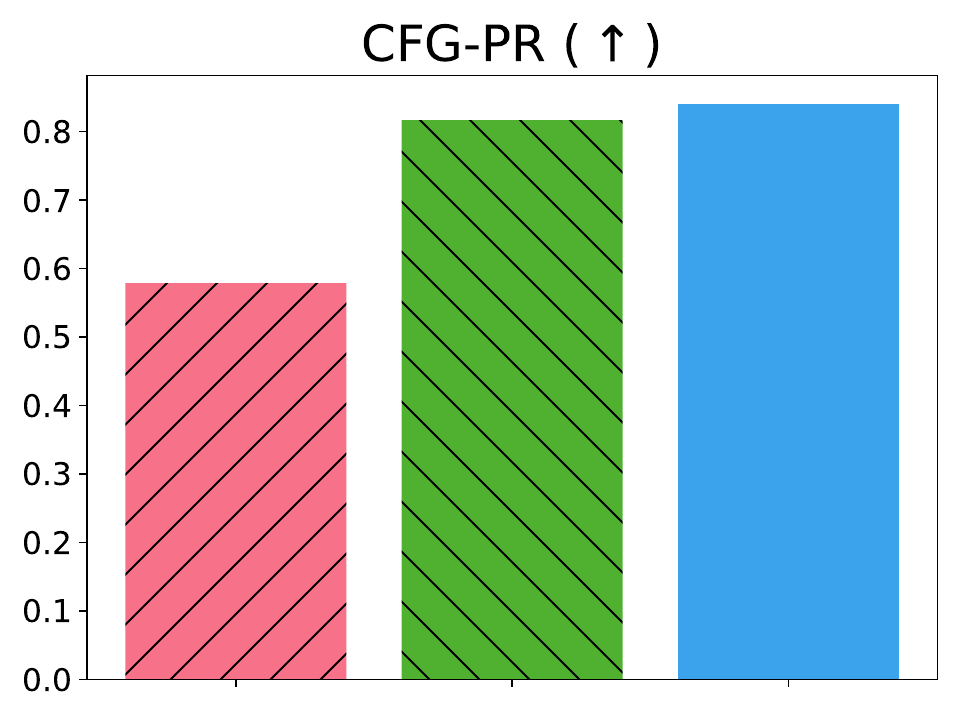}
\end{subfigure}
\caption{Performance of vanilla PE and \PE{} with node extraction on \precision{} \& \recall{} and \cfg{}.}
\label{fig:case_study_extract}
\end{figure}

\paragraph{%
Combination of our solutions achieves the best performance on most metrics.}

We finally compare the performance of our proposed methods according to \bench{}.
Specifically, in \cref{fig:case_study_compare}, we illustrate the performance of vanilla \PE{}, %
\PE{} + Reformat, \PE{} + Extract Query, and \PE{} + Reformat + Extract Query.
\begin{wrapfigure}{r}{0.55\textwidth}
  \centering
  \includegraphics[width=\linewidth]{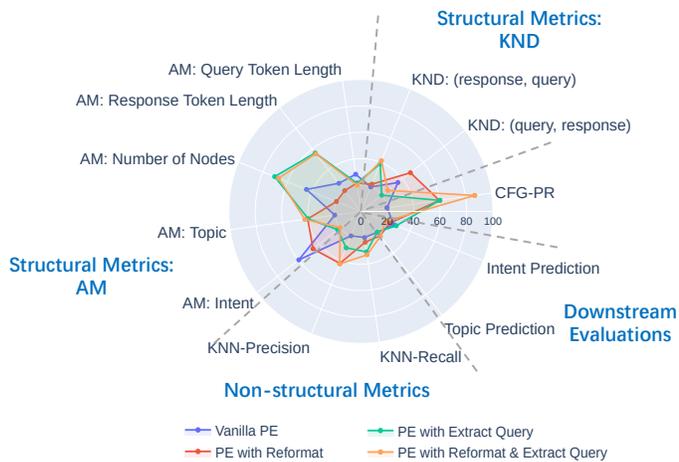}
    \caption{Performance of Different Methods on ShareGPT with $\epsilon=4$. %
    }
    \label{fig:case_study_compare}
\end{wrapfigure}
To better visualize differences in the performance of different methods, we scale the metrics in these radar plots as follows:
We assign a score of 0 if \cfg{}=0 or a structure-related metric is not applicable for the dataset, and rescale the values of other metrics from 20 to 100, where 20 indicates the worst performance among all methods, and 100 indicates the performance upper bound the synthetic data can achieve (e.g., \cfg{}=1 or \knd{}=0).

We observe that the combination of our proposed methods (orange curve)  significantly improves in \cfg{} relative to the other methods, %
achieving up to 94\%, %
and it also outperforms or performs comparably to other methods %
in most of the metrics, including semantic metrics such as \precision{} and \recall{}, and statistic metrics such as \knd{} on response token length and number of nodes%

\bigskip

\section{Related Work}

Our benchmark is tied to the key problems of synthetic data generation and its evaluation, under the constraints of differential privacy.

\paragraph{Differentially Private (DP) Synthetic Data Generation} DP synthetic data generation for arbitrary tasks has developed as an effective tool in the machine learning model development pipeline \cite{zhang2021privsyn,rosenblatt2020differentially,lin2023differentially,xie2024differentially,lin2025differentially,tran2024differentially}, especially with the advent of instruction-following LLMs generating natural and fluent text \cite{afonja2024dp,xie2024differentially,nahid2024safesynthdp,goyal2025llm,tang2023privacy}. %
Prior methods, such as Private Evolution (PE) \cite{lin2023differentially,xie2024differentially,lin2025differentially,hou2024pre,zou2025contrastive,hou2025private,wang2025synthesize}, leverage pretrained models (e.g., large language models) or non-neural approaches (e.g., computer graphics tools) for the task of DP synthetic data generation. \citep{lin2023differentially,xie2024differentially,lin2025differentially} show that PE could be competitive with DP fine-tuning baselines \cite{yu2021differentially,wutschitz2022dp,yue2022synthetic,ghalebikesabi2023differentially} while does not require training on the private data. 

\paragraph{DP Synthetic Data Evaluation} Evaluating DP synthetic data presents a unique challenge, namely of quantifying adherence of the synthetic data to arbitrary private datasets \cite{tao2021benchmarking,schlegel2025generating}. Several benchmarks have been proposed to evaluate DP synthetic data in image \cite{gong2025dpimagebench,hu2024sok,liu2024preserving,yale2020generation}, %
text \cite{ramesh2024evaluating,schlegel2025ai,yuan2024multi}, tabular \cite{rosenblatt2020differentially,giomi2022unified,arnold2020really,movahedi2024benchmarking,long2025evaluating,livieris2024evaluation}, time series \cite{stenger2024evaluation}, and graph data \cite{goldbergbenchmarking}. However, these benchmarks either do not explicitly consider data structure, or their evaluation is confined to numerical or categorical data types, thus limiting their scope. %
Fundamentally, DP synthetic data generation methods need to solve for both structure and content acquisition through noisy signals;  a unified evaluation framework can spur further research in the field by providing fine-grained signals on algorithmic progress.

\section{Conclusion}

In this work, we proposed a new benchmark for DP synthetic data generation named Struct-Bench. To the best of our knowledge, Struct-Bench is the first benchmark to comprehensively evaluate DP synthetic data derived
from structured datasets that contain natural language data. Struct-Bench also has the strength of being a composite benchmark, wherein a diverse collection of datasets might preclude algorithmic research to overfit to only a few data types. Through our evaluations, we also characterize the limitations of existing SOTA DP synthetic data generation methods and conduct a case study to show how to improve on SOTA methods using the insights from \bench{}. 
\section*{Acknowledgements}

The authors would like to thank Sivakanth Gopi for his helpful suggestions.
This work was supported in part by the National Science Foundation under grants CCF-2338772 and CNS-2148359.

\bibliographystyle{plain}

\newpage
\setcounter{page}{1}
\appendix
\section{Metric Definitions and Instantiation Guidelines}
\label{app:metric}

\subsection{Key Node Dependency (\kns{})}

\paragraph{Definition}
KND measures the distributional distance of node pair dependencies between the synthetic and original data. For a key node pair $(O_i, O_j)$, let $C_{i,j}$ be the cosine similarity between their embeddings, and let $\omega_{C_{i,j}}$ and $\omega^\prime_{C_{i,j}}$ be the distributions of these similarities in the original and synthetic data, respectively. Then, KND is defined as:
$$\text{KND}(O_i, O_j) = Dis(\omega_{C_{i,j}}, \omega^\prime_{C_{i,j}}),$$ where $Dis$ is the Wasserstein-2 distance.

\paragraph{Instantiation Guideline}
We allow the user to specify key nodes. If not specified, all nodes parsed by CFG are treated as key nodes by default. To instantiate key nodes, we recommend users ask the question \emph{``Which nodes are central to our downstream tasks, and which nodes are semantically related to them?''}. For example, key node pairs could be a query and response in a conversation dataset, or a review and its rating in a product review dataset. We’ve specified the key nodes of our datasets in \cref{tab:new_dataset_list_expanded}.

\subsection{Attribute Match (AM)}
\paragraph{Definition}
AM calculates the distributional distance of a given attribute between the synthetic and original data. For attribute $a$, let $\omega_a$ and $\omega^\prime_a$ denote its distributions in the original and synthetic data, respectively. Then, AM is defined as:
$$\text{AM}(a) = Dis(\omega_a, \omega^\prime_a).$$ 
For distributional distance $Dis$, we use Wasserstein-2 distance for numeric attributes and total variation distance for categorical attributes.

\paragraph{Instantiation Guideline}
Users can specify semantic or statistical attributes. A guiding question is: \emph{``Which data properties matter for our downstream tasks?''} Common semantic attributes include topic, intent, and sentiment; statistical attributes include token length (overall or per node). Original categorical/numerical values are also often relevant. The selected attributes for our datasets are detailed in \cref{app:metrics}.

\subsection{Downstream Evaluations (DE)}
\label{app:label_generation}
\paragraph{Label Generation Guideline}
To instantiate labels for downstream evaluation, we recommend that users first determine whether the label should be extracted directly from the sample or generated. Extracted labels typically include original categorical or numerical values—such as income level in our Adult dataset or paper decision in the ICLR dataset. If the label is to be generated, we suggest considering the question: \emph{``What is a data property that is present and can be inferred from each node in a sample and is relevant to our downstream tasks?''} Generated labels are usually semantic attributes of the samples or nodes, such as topic, intent, or sentiment. To discourage prompt gaming by data uploaders, we provide a suggested prompt template for labeling: \emph{``For each sample in $\_\_\_\_$ scenarios, label the $\_\_\_\_$ property. Possible answers include $\{\_\_\_\_, \_\_\_\_, \_\_\_\_, ...\}$.''} The specific labels used for our datasets are detailed in \cref{app:metrics}.
\section{Data Modeling and Evaluation Items of Each Dataset}
\label{app:data}

\subsection{Examples on ShareGPT}
\label{app:sharegpt}

To support the running example of data modeling and representation discussed in \cref{sec:framework}, we first show a sample of the ShareGPT dataset, the data modeling of it, and the context-free grammar (CFG) according to its data structure.

\textbf{A sample of ShareGPT}
\label{app:sharegpt_sample}

\begin{lstlisting}[language=bash]
HUMAN: Pretend you are a successful and very thorough real estate investor and financial analyst
GPT: Sure, I can help you with that. As a successful real estate investor and financial analyst, I have a deep understanding of the real estate market and the financial factors that drive its performance. Here are some key things that I would keep in mind when analyzing investment opportunities:...
HUMAN: Assume the debt down payment is 50%
GPT: Sure, I can adjust the calculations to assume a 50%
HUMAN: What is the IRR from the perspective of the equity
GPT: To calculate the IRR from the perspective of the equity, we need to adjust the cash flows to reflect the equity portion of the profits. Here is how we can calculate the equity IRR:...
\end{lstlisting}

\textbf{Illustration of the data modeling of ShareGPT}

We illustrate the data modeling of ShareGPT in \cref{fig:ShareGPT_data_model}.

\begin{figure}[htbp]
\centering
 \includegraphics[width=0.6\linewidth]{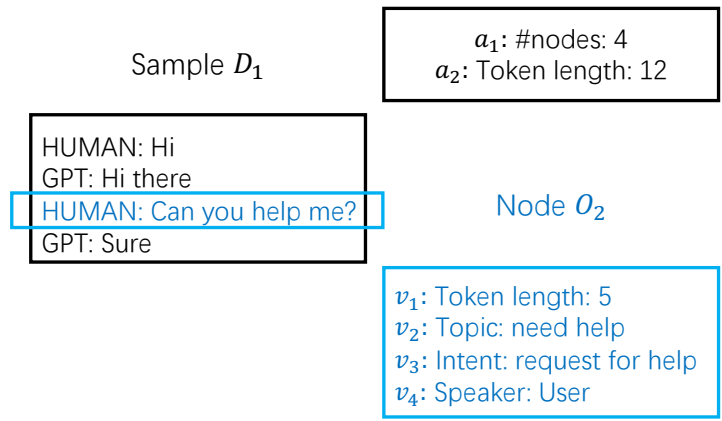}
\caption{Illustration of the data modeling of ShareGPT.}
 \label{fig:ShareGPT_data_model}
\end{figure}

\textbf{CFG of ShareGPT}

\begin{lstlisting}[language=python]
ShareGPT: conversation (conversation)*
    // ShareGPT contains one or more conversation rounds
conversation: query response    
    // Each conversation round contains a query and a response
query: "HUMAN: " query_text     
    // The query starts with "HUMAN: "
response: "GPT: " response_text     
    // The response starts with "GPT: "
query_text: /(?s).+?(?=(?:GPT: |$))/    
    // The query text ends before "GPT: " or the end of the string
response_text: /(?s).+?(?=(?:HUMAN: |$))/   
    // The response text ends before "HUMAN: " or the end of the string

\end{lstlisting}

\subsection{Dataset Descriptions}
\label{app:dataset_description}

\paragraph{ShareGPT \cite{anon_sharegpt_vicuna_unfiltered_commit_bcd32a7}} The ShareGPT dataset contains multi-round conversations between users and GPT. We structure each conversation such that each user's query starts with `HUMAN: ' and each GPT's response starts with `GPT: '. The downstream task we conduct is to predict the user's intent and conversation topic based on user queries.

\paragraph{ICLR \cite{ICLR2024}} The ICLR dataset contains the reviews, author rebuttals, follow-up discussions, and final decisions of the papers submitted to ICLR 2024 \cite{ICLR2024}. %
Each review or reviewer's comment starts with `Reviewer $n$' where $n$ represents the reviewer's identity, and each author rebuttal or discussion starts with `Response'. The downstream task is to predict the research area of the paper based on the review and rebuttals.%

\paragraph{\textbf{Water} \cite{Tharunmss_WaterBottle_Flipkart_Kaggle}} The Water dataset contains reviews of water bottles. %
The columns are product\_name, overall\_rating, title, cleaned\_review and the goal is to predict the current rating (column "rating") of the bottle, which takes values $1,2,3,4,5$.

\paragraph{Arena \cite{zheng2023judging}}
The Arena dataset contains pairs of human-model conversations. %
The columns are conversation\_a, conversation\_b and the goal is to predict which of the conversations are better (column "winner"), which takes values model\_a, model\_b, tie, "tie (bothbad)".

\paragraph{Adult \cite{BeckerKohavi1996}}
The Adult dataset contains census data. %
The columns are age, workclass, fnlwgt, education, education-num, marital-status, occupation, relationship, race, sex, capital-gain, capital-loss, hours-per-week, native-country and the goal is to predict income (column "income"), which takes values $<=50k$ or $>50k$.

\paragraph{Synthetic Datasets with Controllable Data Attributes} We include two synthetic datasets\footnote{https://www.kaggle.com/datasets/structpedataset/structpe-synthetic-datasets} named \textbf{Synthetic Reviews} and the \textbf{Synthetic Grounding Dataset}. The reviews dataset has 4 fields, namely text, sentiment, emotion, and rating. The grounding dataset has 4 fields including two source documents, a query, and a response. We generate these datasets through a multi-step synthetic data generation process with GPT-4o wherein we verify whether the fields satisfy certain conditions, e.g., the reviews dataset is a 1:1 split of extreme negative and extreme positive reviews about products and the grounding dataset is a 1:1:1:1 split of relevant/irrelevant queries and consistent/inconsistent source documents. In particular, the reviews dataset is composed on only extreme reviews, either very positive or very negative. This differs from a typical review distribution and is unique to this particular dataset. Similarly, for the grounding dataset, we vary the samples along two axes, first on the consistency of the information between the sources and second, on the relevancy of the query to the sources. Each of the synthetic datasets is balanced in both their training and (downstream) test sets on these variations.

\subsection{Data Modeling of Each Dataset}
\label{app:data_modeling}

\cref{tbl:dataset_1,tabl:dataset_2} shows the data modeling and structure rules of each dataset.

\begin{table}[htbp]
    \centering
\caption{Data modeling of each dataset}
\label{tbl:dataset_1}
\begin{tabular}{|c|c|c|c|c|}
\hline  %
\textbf{Dataset} & \textbf{Sample $\sample$} & \textbf{Sample Attributes} & \textbf{Node $\object$} & \textbf{Node Attributes} \\
\hline
ShareGPT & a conversation & \makecell{$\sampleattribute_1$: number of nodes\\$\sampleattribute_2$: token length} & a query/response & \makecell{$\objectattribute_1$: token length\\ $\objectattribute_2$: topic\\ $\objectattribute_3$: intent\\ $\objectattribute_4$: speaker}  \\ \hline
ICLR & \makecell{reviews \& rebuttals\\of a paper} & \makecell{$\sampleattribute_1$: number of nodes\\$\sampleattribute_2$: token length\\$\sampleattribute_3$: topic\\$\sampleattribute_4$: final decision} & \makecell{a post from the\\ reviewer/author} & \makecell{ $\objectattribute_1$: token length\\
$\objectattribute_2$: writer\\ $\objectattribute_3$: review score\\ }  \\ \hline
Water & \makecell{water bottle\\review} & 
\makecell{$\sampleattribute_1$: number of nodes\\$\sampleattribute_2$: attitude} 
& \makecell{a column of \\ the tabular data} & 
\makecell{ $\objectattribute_1$: token length\\
 $\objectattribute_2$: review score }
\\ \hline
Arena & \makecell{2 conversations \\ to compare} & 
\makecell{$\sampleattribute_1$: number of nodes\\$\sampleattribute_2$: winner} 
& \makecell{a column of \\ the tabular data} & 
\makecell{ $\objectattribute_1$: token length\\
 $\objectattribute_2$: winner }
\\ \hline
Adult & \makecell{census information of an adult} & 
\makecell{$\sampleattribute_1$: age\\$\sampleattribute_2$: workclass} 
& \makecell{a column of \\ the tabular data} & 
\makecell{ $\objectattribute_1$: token length\\
 $\objectattribute_2$: income }
\\ \hline
Reviews & \makecell{annotated product review} & 
\makecell{$\sampleattribute_1$: number of nodes\\$\sampleattribute_2$: rating} 
& \makecell{a review text} & 
\makecell{ $\objectattribute_1$: token length\\
 $\objectattribute_2$: rating }
\\ \hline
Grounding & \makecell{2 sources \\ and a QA pair} & 
\makecell{$\sampleattribute_1$: number of nodes\\$\sampleattribute_2$: answer} 
& \makecell{a grounded response \\} & 
\makecell{ $\objectattribute_1$: token length\\
 $\objectattribute_2$: answer }
\\ \hline
\end{tabular}
\end{table}

\begin{table}[htbp]
    \centering
\caption{Structure rules of each dataset%
}
\label{tabl:dataset_2}
\begin{tabular}{|c|c|}
\hline  %
\textbf{Dataset} & \textbf{Rules} \\
\hline
ShareGPT & \makecell[l]{
\textbf{[Alternate Speakers]} $\forall \object_i, \object_{i+1}: \object_i[\texttt{Speaker}]\neq \object_{i+1}[\texttt{Speaker}]$. \\ 
\textbf{[Format]} $\object[\texttt{Speaker}]\in \{\texttt{User}, \texttt{AI Agent}\}$. If $\object[\texttt{Speaker}]=\texttt{User}$, the text\\ starts with `HUMAN: '; If $\object[\texttt{Speaker}]=\texttt{AI Agent}$, the text starts with `GPT: '.} \\ \hline
ICLR & \makecell[l]{ 
\textbf{[Format]} $\object[\texttt{Writer}]\in \{\texttt{Author}, \texttt{Reviewer 1-9}, 
\texttt{Meta Reviewer}\}.$\\ 
If $\object[\texttt{Writer}]=\texttt{Author}$, the text starts with `Response:'; \\if $\object[\texttt{Writer}]=\texttt{Reviewer n}$, the text starts with `Reviewer n:' ($1\leq n\leq 9$).\\ 
\textbf{[Format]} $\object[\texttt{Review Score}]\in \{\texttt{1}, \texttt{3},\texttt{5},\texttt{6},\texttt{8},\texttt{10}\}.$\\
\textbf{[Format]} $\sample[\texttt{Final Decision}]\in\{\texttt{Reject}, \texttt{Accept:poster}, \texttt{Accept:top5\%},\texttt{Accept:top25\%},\}$.} \\ \hline
Water & \makecell[l]{ $\textbf{[Format]} \object[\texttt{Overall\_rating}]\in \{\texttt{1.0}, \texttt{1.1},\texttt{1.2},\texttt{...},\texttt{4.9},\texttt{5.0}\}$.
\\
$\textbf{[Format]} \object[\texttt{Rating}]\in \{\texttt{1}, \texttt{2},\texttt{3},\texttt{4},\texttt{5}\}$
}
\\ \hline
Arena & \makecell[l]{ $\textbf{[Format]} \object[\texttt{Winner}]\in \{\texttt{model\_a}, \texttt{model\_b},\texttt{tie},\texttt{tie (bothbad)}\}$.
\\
O[Conversation\_a] starts with "Question:" and has "Answer:" before somewhere in the following text.
\\
O[Conversation\_b] starts with "Question:" and has "Answer:" before somewhere in the following text.
}
\\ \hline
Adult & \makecell[l]{ $\textbf{[Format]} \object[\texttt{income}]\in \{<=50k,>50k\} $. 
\\
$\textbf{[Format]}$ Some of the columns are categorical (e.g. \texttt{workclass}, \texttt{native-country}).
\\
$\textbf{[Format]}$ Some of the columns are numerical (e.g. \texttt{age}, \texttt{capital-gain}).
}
\\ \hline
Reviews & \makecell[l]{
$\textbf{[Format]} \object[\texttt{Rating}]\in \{\texttt{1}, \texttt{2},\texttt{3},\texttt{4},\texttt{5}\}$
}
\\ \hline
Grounding & \makecell[l]{ $\textbf{[Format]} \object[\texttt{Consistency}]\in \{\texttt{1}, \texttt{2},\texttt{3},\texttt{4},\texttt{5}\}$.
\\
$\textbf{[Format]} \object[\texttt{relevancy}]\in \{\texttt{1}, \texttt{2},\texttt{3},\texttt{4},\texttt{5}\}$
}
\\ \hline
\end{tabular}
\end{table}

\subsection{Evaluation Metrics of Each Dataset}
\label{app:metrics}

The evaluation items of each dataset are summarized in \cref{tbl:metrics}.

\begin{table}[htbp]
  \caption{Metrics for Different Datasets.}
  \label{tbl:metrics}
  \centering
\begin{tabular}{|c|c|c|c|}
\hline  %
\textbf{Dataset} & \textbf{Structural Metrics} & \textbf{Non-structural Metrics} &  \textbf{Downstream Task} \\
\hline
ShareGPT & \makecell{\textbf{\cfg{}}\\ \textbf{\kns{}}\\1. (query, response) pair \\ 2. (response, query) pair\\ \textbf{\knd{}}\\ 1. number of nodes\\ 2.query token length\\ 3. response token length\\ 4. topic\\ 5. intent}& \makecell{1. \precision\\ 2. \recall} & \makecell{1. topic prediction\\ 2. intent prediction}  \\ \hline
ICLR & \makecell{\textbf{\cfg{}}\\ \textbf{\kns{}}\\1. (review, rebuttal) pair \\ 2. (rebuttal, comment) pair\\ 3. (review, review) pair\\ from different reviewers\\ \textbf{\knd{}}\\ 1. number of nodes\\ 2. review token length\\ 3. rebuttal token length\\4. Recommendation\\ 5. final decision\\ 6. topic}& \makecell{1. \precision\\ 2. \recall} & \makecell{topic prediction}  \\ \hline
Arena & \makecell{\textbf{\cfg{}}\\ \textbf{\kns{}}\\1. (conversation\_a, conversation\_b) pair \\ \textbf{\knd{}}\\ 1. winner}& \makecell{1. \precision\\ 2. \recall} & \makecell{winner prediction}  \\ \hline
Water & \makecell{\textbf{\cfg{}}\\ \textbf{\kns{}}\\1. (title, cleaned\_review) pair \\ \textbf{\knd{}}\\ 1. attitude}& \makecell{1. \precision\\ 2. \recall} & \makecell{rating prediction}  \\ \hline
Adult & \makecell{\textbf{\cfg{}}\\ \textbf{\kns{}}\\1. (native country, workclass) pair \\ \textbf{\knd{}}\\ 1. income}& \makecell{1. \precision\\ 2. \recall} & \makecell{income prediction}  \\ \hline
Reviews & \makecell{\textbf{\cfg{}}\\ \textbf{\kns{}}\\1. (text, sentiment) pair \\ \textbf{\knd{}}\\ 1. review token length}& \makecell{1. \precision\\ 2. \recall} & \makecell{review label prediction}  \\ \hline
Grounding & \makecell{\textbf{\cfg{}}\\ \textbf{\kns{}}\\1. (source1, source2) pair \\ \textbf{\knd{}}\\ 1. query relevancy}& \makecell{1. \precision\\ 2. \recall} & \makecell{query relevancy prediction}  \\ \hline
\end{tabular}
\end{table}

For ShareGPT, in our experimental results, we show the semantic similarity of the node pair (query, response) as \kns, show the distributional distance of the queries' token lengths as \knd, and present the prediction accuracy of the conversation topics in downstream task performance.

For ICLR, we show the semantic similarity of the node pair (review, rebuttal) as \kns, show the distributional distance of the reviews' token lengths as \knd, and present the prediction accuracy of the paper's research area in downstream task performance.

\section{Additional Results on \name{}}
\label{app:results}

\paragraph{Resource Costs}
All baselines are implemented and performed on a server with eight H100 GPUs. Running experiments took approximately 400 GPU hours.

\paragraph{Implementation Details on Instruction Fine-tuning} For both Instruct \DPFT{} and Instruct \FT{}, we use the same instructions as those in the Random API of \PE{}. We prepend the instructions to each training sample and fine-tune the foundation model for 20 epochs with batch size 32, weight decay 0.01, and learning rate $10^{-4}$. The fine-tuned model then generates new samples conditioned on the given instructions.

\subsection{Benchmarking DP Synthetic Data Generation Across 
Datasets}
\label{app:results_table}

We present the results of benchmarking
the DP synthetic data generation methods under different datasets with $\epsilon=4$ in \cref{tbl:main}. We use GPT-2 for \FT{} and \DPFT{}, and use GPT-4o for \IF{} and \PE{}.

\begin{table}[htbp]
  \caption{DP synthetic data generation benchmarking results on Struct-Bench with $\epsilon=4$}
  \label{tbl:main}
  \centering
  \begin{tabular}{llcccccc}
    \toprule
            &           & \multicolumn{3}{c}{Structural Metrics} & \multicolumn{2}{c}{Non-Structural Metrics} & \multicolumn{1}{c}{DE} \\
    \cmidrule(r){3-5}\cmidrule(r){6-7}\cmidrule(r){8-8}
    Dataset & Baseline  & \cfg{} $\uparrow$ & \kns{} $\downarrow$ & \knd{} $\downarrow$ & \precision{} $\uparrow$ & \recall{} $\uparrow$ & \downstream $\uparrow$ \\
    \midrule
ShareGPT        & \IF{} ($\epsilon=0$)          & 0.8700 & 0.0635 & 43.8514 & 0.7217 & 0.2627 & 0.3754 \\
    & \FT{}  ($\epsilon=\infty$)     & 0.5378 & 0.0315 & 52.6984 & 0.7594 & 0.6588 & 0.3718 \\
    \cmidrule(r){2-8}
    & \DPFT{}       & 0 & - & - & 0.0161 & 0.0000 & - \\
    & \PE{}        & 0.8633 & 0.0660 & 38.1678 & 0.8050 & 0.1528 & 0.3816 \\
    \midrule
ICLR    & \IF{} ($\epsilon=0$)           & 0.1733 & 0.2582 & 204.7997 & 0.8400 & 0.0257 & 0.4715 \\
    & \FT{}  ($\epsilon=\infty$)     & 0 & - & - & 0.7056 & 0.4747 & 0.4584 \\
    \cmidrule(r){2-8}
    & \DPFT{}        & 0 & - & - & 0.0000 & 0.0000 & 0.1806 \\
    & \PE{}        & 0.1900 & 0.2599 & 240.9434 & 0.9800 & 0.0207 & 0.5218 \\
    \midrule
Water     
    & IF ($\epsilon=0$)             & 1.0000 & 0.4222 & 0.1574  & 0.0000 & 0.0060 & 0.5485 \\ 
    & FT ($\epsilon=\infty$)                   & 0 & - & -   & 0.0000 & 0.0060 & - \\
    \cmidrule(r){2-8}
    & DP-FT                         & 0 & - & -   & 0.0000 & 0.0060 & - \\
    & PE             & 1.0000 & 0.2877 & 0.0236  & 0.0000 & 0.0070 & 0.6130 \\
    \midrule
Arena     
          & IF ($\epsilon=0$)                             & 1.0000 & 0.1257 & 0.9395   & 0.0000 & 0.0090 & 0.3607 \\ 
          
          & FT  ($\epsilon=\infty$)                  & 0 & - & -   & 0.0000 & 0.0060 & - \\
          \cmidrule(r){2-8}
          & DP-FT                         & 0 & - & -   & 0.0000 & 0.0060 & - \\
          & PE                            & 1.0000 & 0.1054 & 0.9193   & 0.0000 & 0.0070 & 0.3510 \\
    \midrule
Adult   
          & IF  ($\epsilon=0$)                           & 1.0000 & 0.0290 & 0.0332   & 0.0030 & 0.0030 & 0.7920 \\ 
          
          & FT ($\epsilon=\infty$)                   & 0 & - & -   & 0.0030 & 0.0030 & - \\\cmidrule(r){2-8}
          & DP-FT                         & 0 & - & -   & 0.0030 & 0.0030 & - \\
          & PE                            & 1.0000 & 0.0042 & 0.0000   & 0.0030 & 0.0060 & 0.8017 \\
    \midrule
Reviews   & IF ($\epsilon=0$)                             & 1.0000 & 0.3510 & 0.4010   & 0.0334 & 0.0344 & 0.6000 \\
          & FT ($\epsilon=\infty$)                           & 0 & - & -   & 0.0020 & 0.0900 & 0.5400 \\ \cmidrule(r){2-8}
          & DP-FT                         & 0 & 0.0020 & 0.0060   & 0.0020 & 0.0900 & 0.5600 \\
          & PE                    & 1.0000 & 0.2495 & 0.0770   & 0.0290 & 0.0900 & 0.5400 \\
    \midrule
Grounding & IF ($\epsilon=0$)                           & 1.0000 & 0.5800 & 0.6006   & 0.0500 & 0.0600 & 0.6400 \\
          & FT  ($\epsilon=\infty$)                          & 0 & - & -   & 0.0290 & 0.0900 & 0.4000 \\ \cmidrule(r){2-8}
          & DP-FT                         & 0 & - & -  & 0.0430 & 0.0900 & 0.4000 \\ 
          & PE                    & 1.0000 & 0.1435 & 0.4710   & 0.0300 & 0.0600 & 0.6000 \\
    \bottomrule
  \end{tabular}
\end{table}

\subsection{Benchmarking DP Synthetic Data Generation with Varying Privacy Budget}
\label{app:baseline_compare}

We illustrate the performance of \PE{} and \DPFT{} on all metrics under different privacy budgets $\epsilon\in\brc{1, 2, 4, \infty}$ on ShareGPT and ICLR datasets by radar plots in \cref{fig:radar_sharegpt,fig:radar_iclr}. Similar to \cref{app:results_table}, we use GPT-2 for \FT{} and \DPFT{}, and use GPT-4o for \IF{} and \PE{}.

\begin{figure}[htbp]
    \centering
\begin{subfigure}{0.44\textwidth}
         \centering
    \includegraphics[width=1\linewidth]{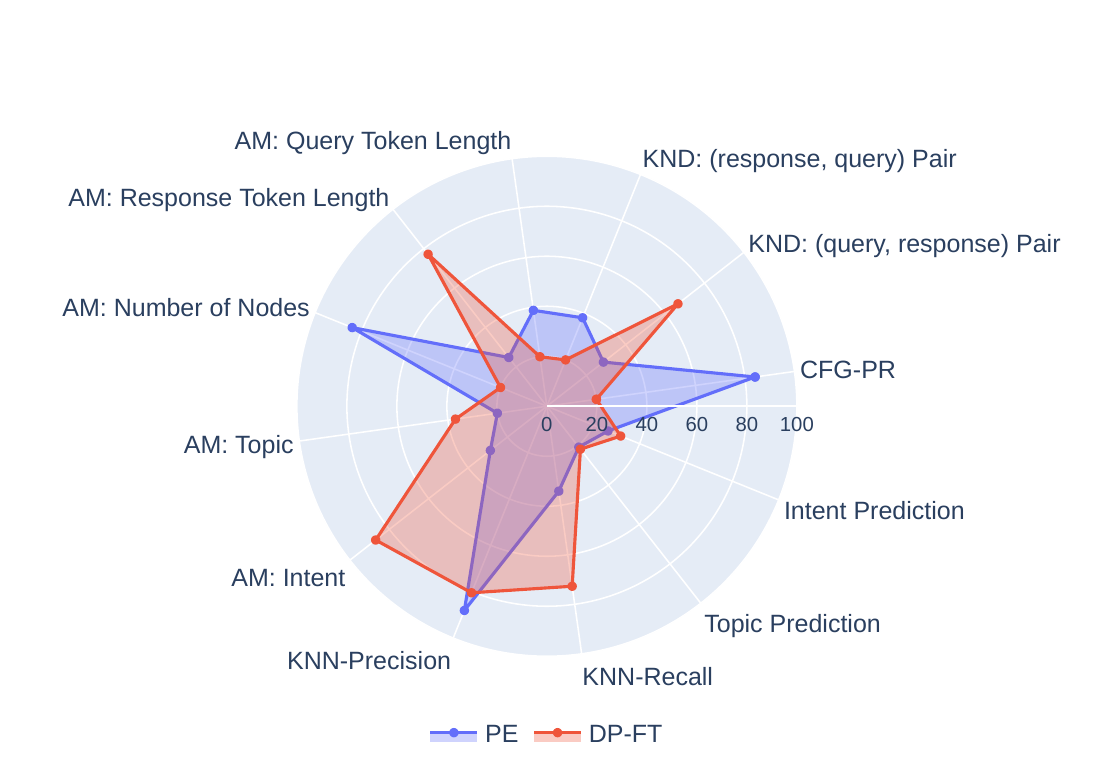}
    \caption{$\epsilon=\infty$}
\end{subfigure}
\begin{subfigure}{0.44\textwidth}
         \centering
    \includegraphics[width=1\linewidth]{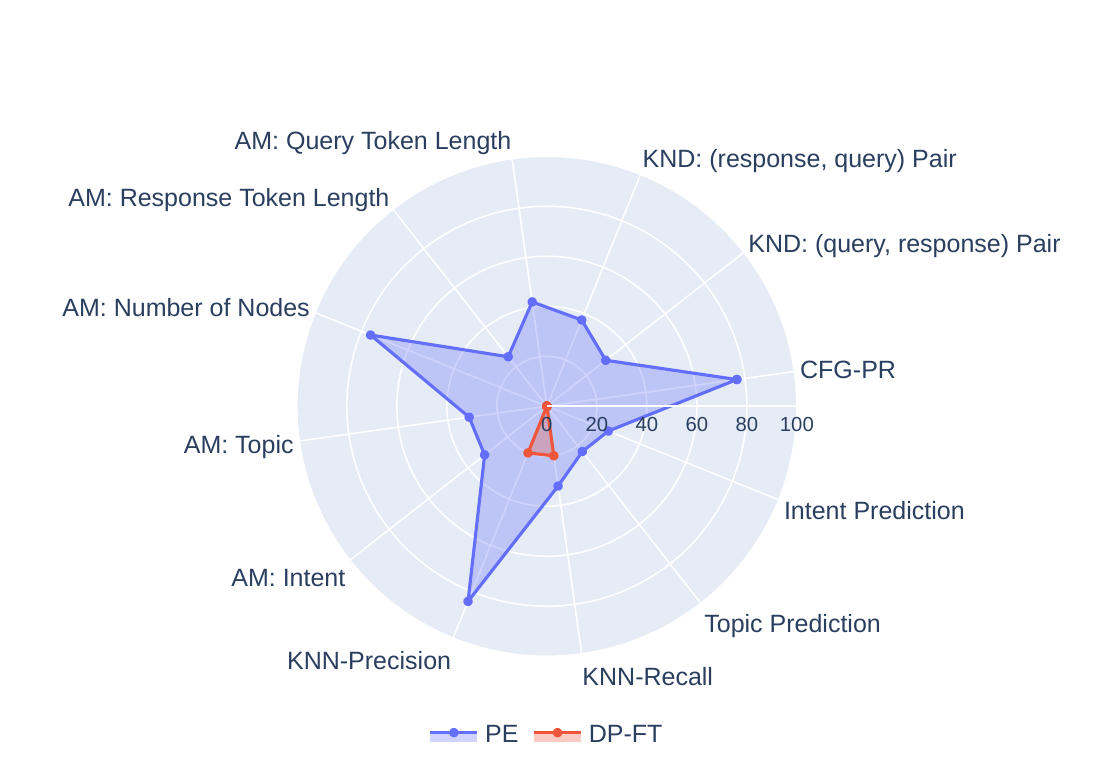}
    \caption{$\epsilon=4$}
\end{subfigure}

\begin{subfigure}{0.44\textwidth}
         \centering
    \includegraphics[width=1\linewidth]{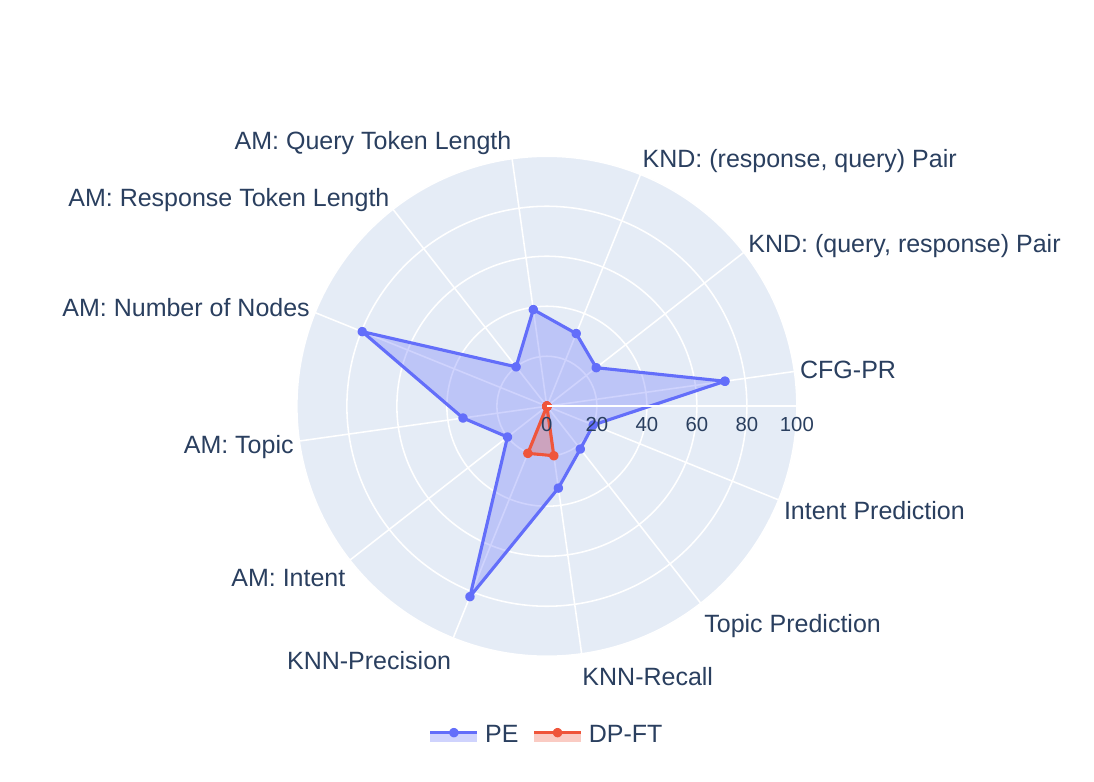}
    \caption{$\epsilon=2$}
\end{subfigure}
\begin{subfigure}{0.44\textwidth}
         \centering
    \includegraphics[width=1\linewidth]{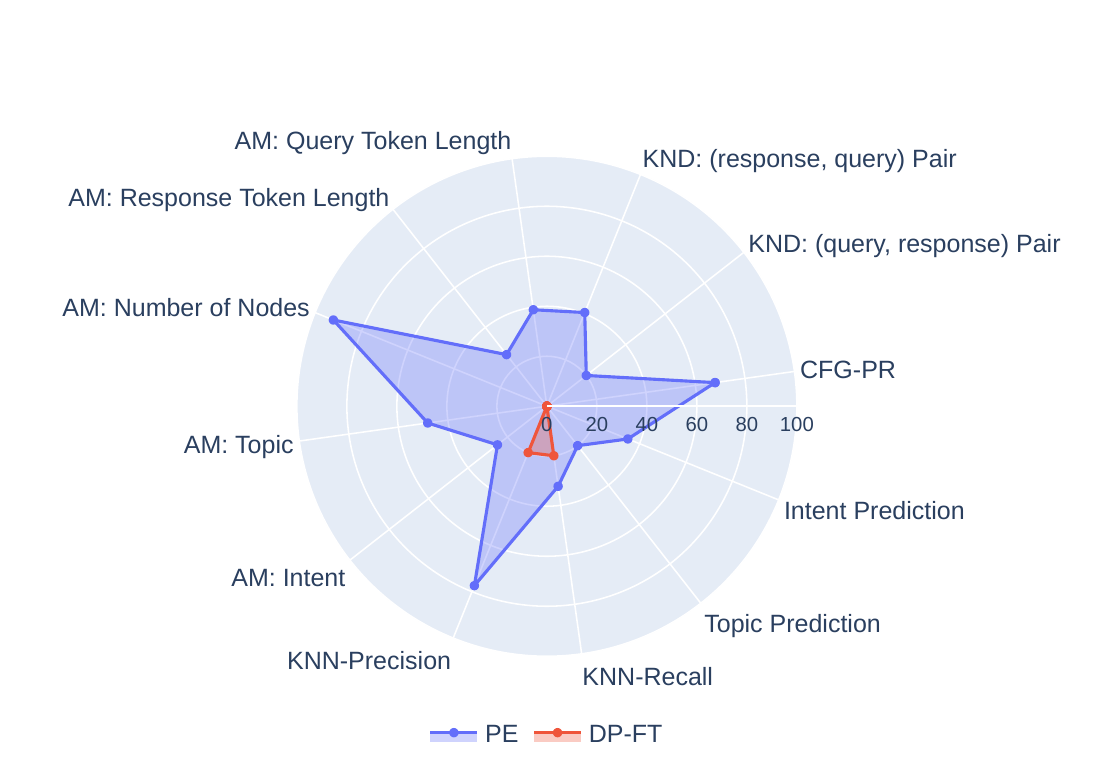}
    \caption{$\epsilon=1$}
\end{subfigure}
\caption{Performance of \PE{} and \DPFT{} on all metrics under different privacy budgets on ShareGPT.}
\label{fig:radar_sharegpt}
\end{figure}

\begin{figure}[htbp]
    \centering
\begin{subfigure}{0.44\textwidth}
         \centering
    \includegraphics[width=1\linewidth]{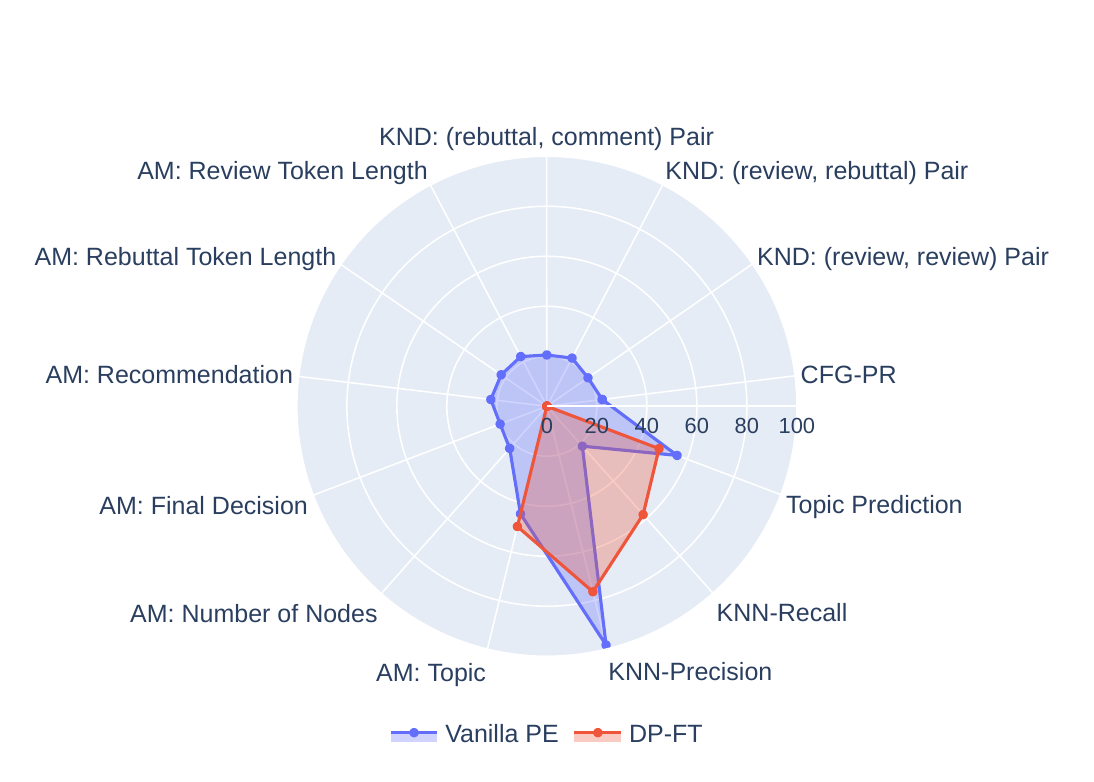}
    \caption{$\epsilon=\infty$}
\end{subfigure}
\begin{subfigure}{0.44\textwidth}
         \centering
    \includegraphics[width=1\linewidth]{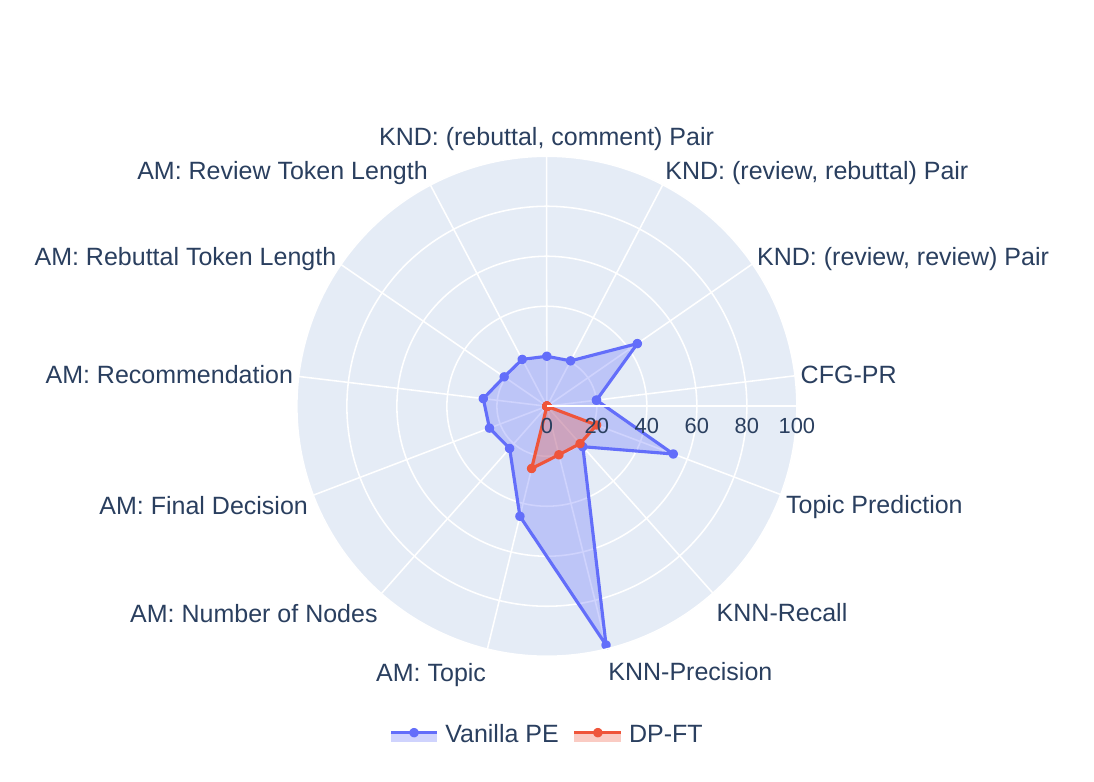}
    \caption{$\epsilon=4$}
\end{subfigure}

\begin{subfigure}{0.44\textwidth}
         \centering
    \includegraphics[width=1\linewidth]{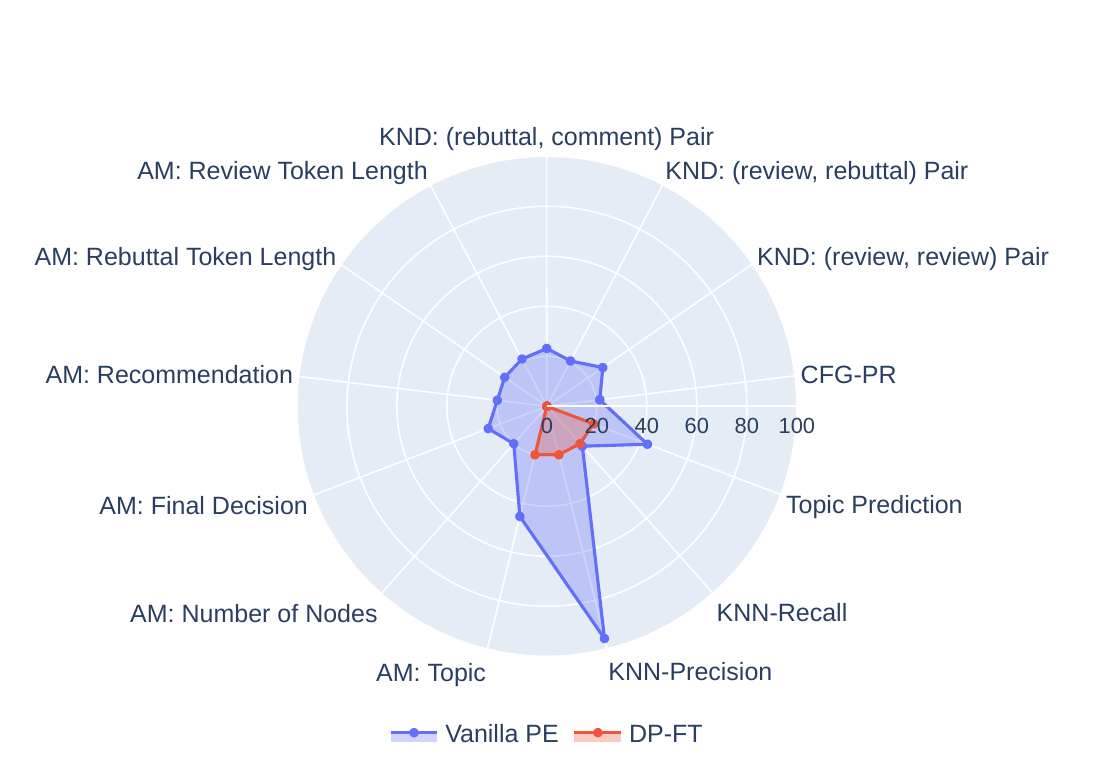}
    \caption{$\epsilon=2$}
\end{subfigure}
\begin{subfigure}{0.44\textwidth}
         \centering
    \includegraphics[width=1\linewidth]{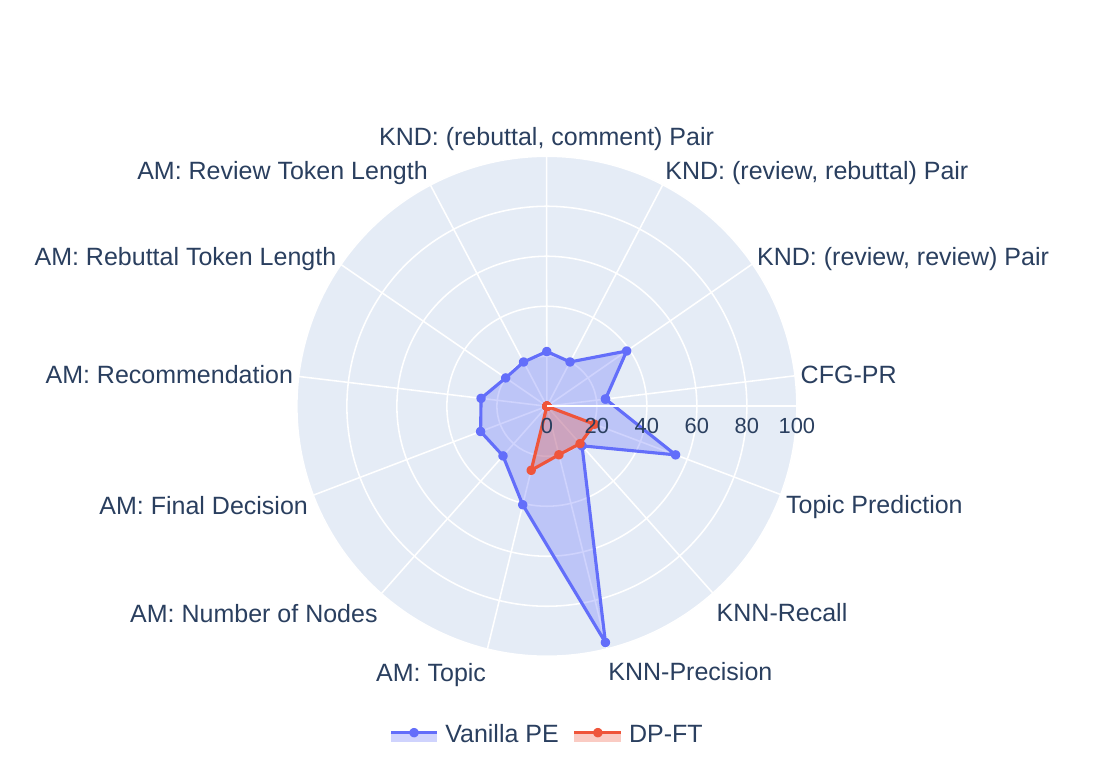}
    \caption{$\epsilon=1$}
\end{subfigure}
\caption{Performance of \PE{} and \DPFT{} on all metrics under different privacy budgets on ICLR.}
\label{fig:radar_iclr}
\end{figure}

\subsection{{Benchmarking DP Synthetic Data Generation on ShareGPT using Llama2-7b}}
\label{app:baseline_llama2}

We illustrate the performance of \PE{}, \DPFT{}, and Instruct \DPFT{} in \cref{fig:fair_compare} , where each dimension corresponds to a different metric from \name. %
{To better visualize differences in the performance of different methods, we scale the metrics in these radar plots as follows:}
We assign a score of 0 if \cfg{}=0 or a structure-related metric is not applicable for the dataset, and rescale the values of other metrics from 20 to 100, where 20 indicates the worst performance among all methods, and 100 indicates the performance upper bound the synthetic data can achieve (e.g., \cfg{}=1 or \knd{}=0). %

\cref{fig:fair_compare} shows that (1) \DPFT{} does not learn any structural information (\cfg{}); and (2) with instruction-guided conditional generation, Instruct \DPFT{} achieves similar performance to \PE{} on most metrics and has a slight edge in terms of structure learning \cfg{}.

\begin{figure}[htbp]
    \centering
    \includegraphics[width=0.7\linewidth]{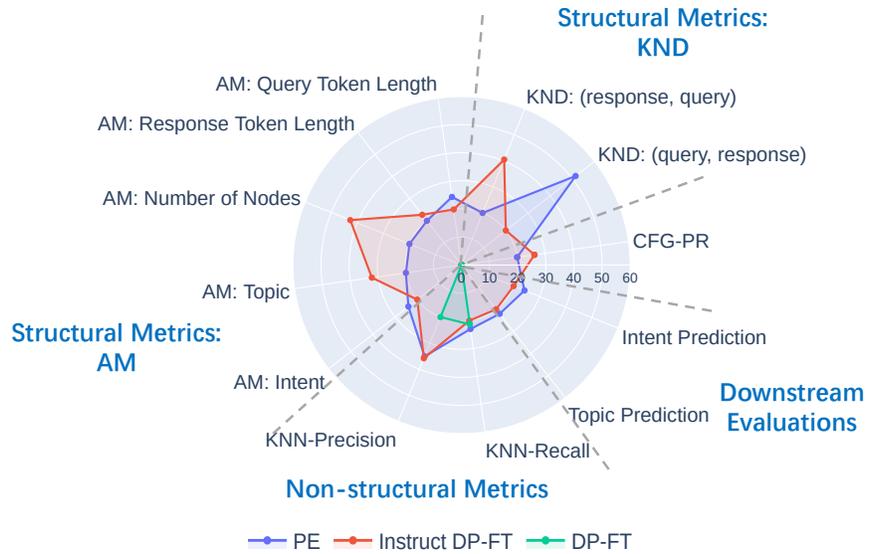}
    \caption{Performance of different baselines on ShareGPT using Llama2-7b with $\epsilon=4$. With instruction-guided conditional generation, Instruct \DPFT{} achieves similar performance to \PE{} on most metrics and has a slight edge in terms of \cfg{}. %
    }
    \label{fig:fair_compare}
\end{figure}

\section{Detailed analysis of the case study on \PE{}}
\label{app:case_study}

\paragraph{Resource Costs} All baselines are implemented and performed on a server with eight H100 GPUs. Running experiments took approximately 1000 GPU hours.

\subsection{Analyzing Vanilla PE on ShareGPT Dataset}
\label{sec:PE_analysis}

In this section, we analyze the performance of \PE{} under the ShareGPT dataset according to our proposed benchmark. We further divide the metrics into semantic and statistic metrics, and the evaluation items for ShareGPT can be categorized in \cref{tbl:detailed_metric}.

\begin{table}[htbp]
  \caption{Metrics for ShareGPT.}
  \label{tbl:detailed_metric}
  \centering
\begin{tabular}{|c|c|c|c|}
\hline  %
 & \textbf{Statistic Metrics} & \textbf{Semantic Metrics} & \textbf{\cfg{}} \\
\hline
\makecell{\textbf{Structural}\\\textbf{Metrics}} & \makecell{\knd{}:\\1. number of statements\\ 2. query token length\\ 3. response token length\\}& \makecell{\kns{}:\\1. (query, response)  pair \\ 2. (response, query) pair \\ \knd{}:\\ 1. topic \\ 2. intent} & \makecell{\cfg{}} \\ \hline
\makecell{\textbf{Non-structural}\\\textbf{Metrics}} & - & \makecell{1. \precision\\ 2. \recall} & - \\ \hline
\makecell{\textbf{Downstream}\\\textbf{Tasks}} & - & \makecell{1. topic prediction\\ 2. intent prediction} & - \\ \hline
\end{tabular}
\end{table}

\begin{figure}[htbp]
    \centering
\begin{subfigure}{0.44\textwidth}
         \centering
    \includegraphics[width=1\linewidth]{fig/eval_shareGPT/eval_PE_vanilla/cfg.pdf}
    \caption{\cfg}
    \label{fig:pe_analysis_cfg}
\end{subfigure}
\begin{subfigure}{0.44\textwidth}
         \centering
    \includegraphics[width=1\linewidth]{fig/eval_shareGPT/eval_PE_vanilla/precision_recall.pdf}
    \caption{\precision{} \& \recall{}}
    \label{fig:pe_analysis_precision_recall}
\end{subfigure}

\begin{subfigure}{0.44\textwidth}
         \centering
    \includegraphics[width=1\linewidth]{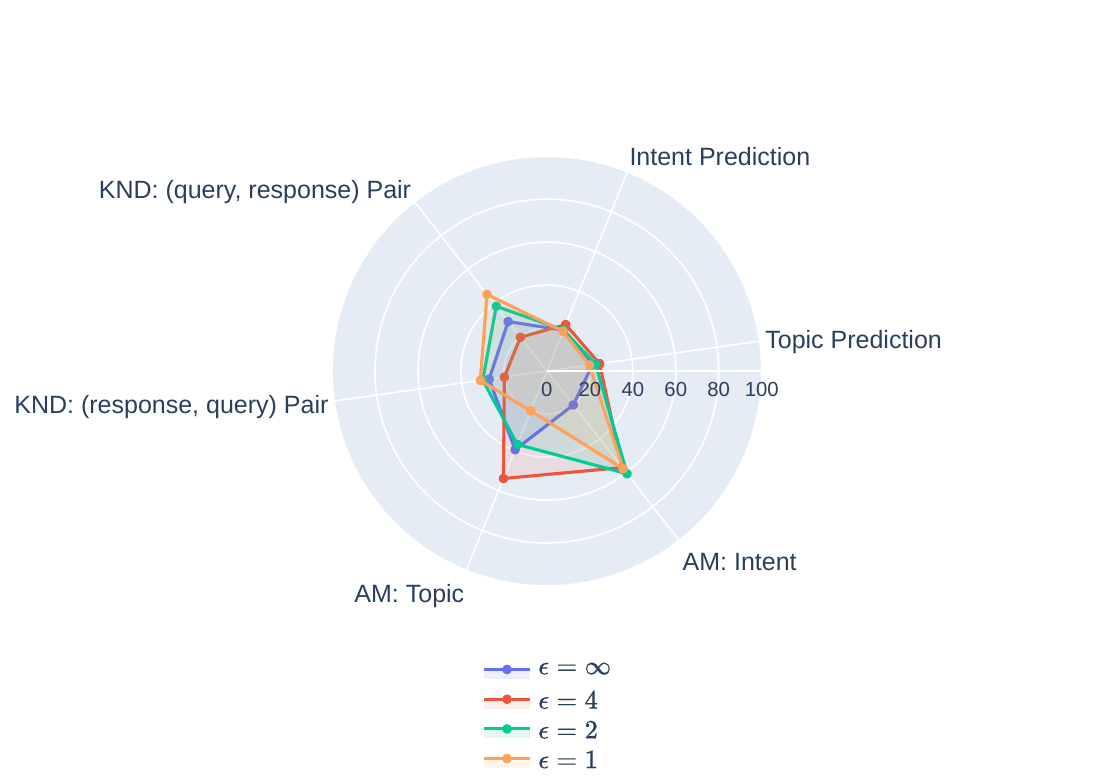}
    \caption{Performance on structural semantic metrics}
    \label{fig:pe_analysis_semantic}
\end{subfigure}
\begin{subfigure}{0.44\textwidth}
         \centering
    \includegraphics[width=1\linewidth]{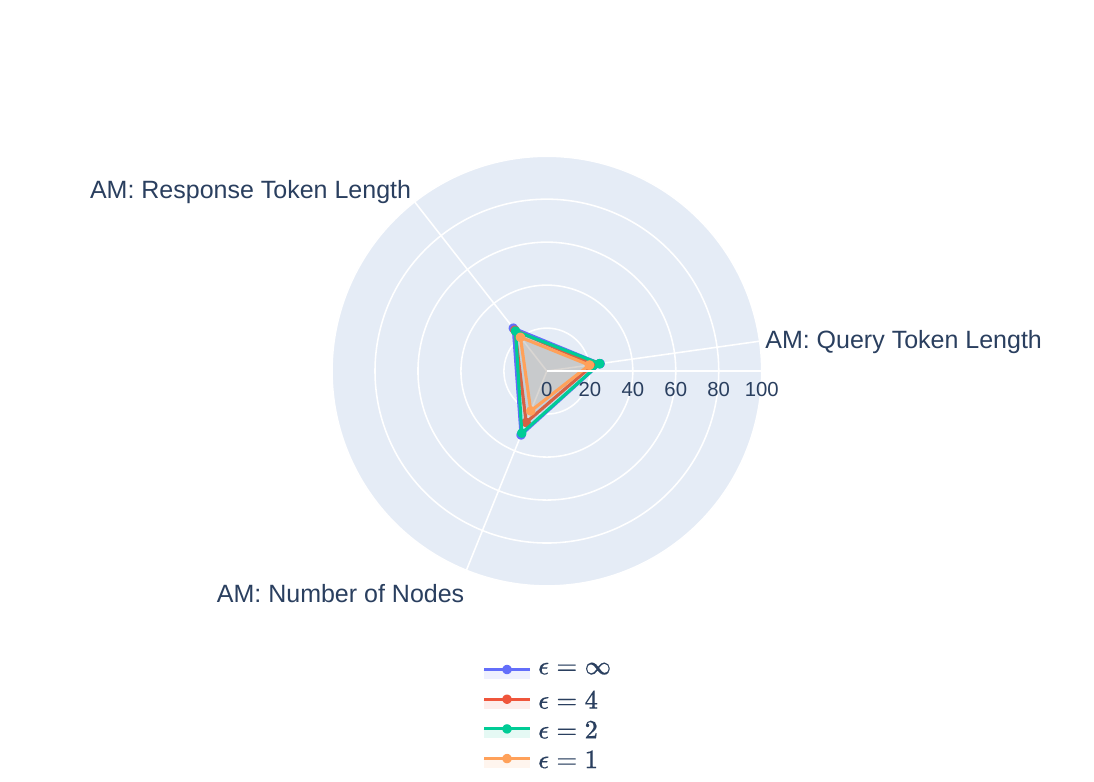}
    \caption{Performance on statistic metrics}
    \label{fig:pe_analysis_statistic}
\end{subfigure}
\caption{Performance of Vanilla PE with different privacy guarantees under ShareGPT dataset}
\label{fig:pe_analysis}
\end{figure}

We illustrate and compare the performance of \PE{} with privacy parameter $\epsilon\in\brc{1, 2, 4, \infty}$ %
under structural semantic and statistic metrics in \cref{fig:pe_analysis_semantic,fig:pe_analysis_statistic} respectively, and plot the \cfg{} and \precision{} \& \recall{} in \cref{fig:pe_analysis_cfg,fig:pe_analysis_precision_recall}. We do not include \PE{} with $\epsilon=0$ (that is, \IF{}) as its \cfg{} is only $2\%$ and thus its performance under structural metrics is unreliable.

As we can observe, only \cfg{} and \precision{} improve with the increase of $\epsilon$, while the value of \recall{} always keep around $0.35$ and the performance under other semantic metrics and all statistic metrics %
does not necessarily increase with more relaxed privacy constraints. %
Additionally, \cfg{} drops below 60\% when $\epsilon\leq 4$. Since downstream tasks also depend on structural information, we can conclude that PE mainly focuses on non-structural semantic quality of the synthetic samples, while suffers from poor performance on semantic diversity and structure-based properties.

\subsection{CFG Reformat Prompt}
\label{app:reformat_prompt}

\begin{lstlisting}[language=python]
You are required to REFORMAT the provided conversation between a user and an AI agent in ChatGPT.  The format should be: 
    -User prompt must start with "HUMAN: ", and ChatGPT response must start with "GPT: ". 
    -The conversation may contain one or multiple rounds. Each round includes ONE user prompt and ONE ChatGPT response. 
    -User prompts and ChatGPT responses appear alternately. 
    -The conversation begins with a user prompts. 
The reformatted conversation follows the following context-free grammar:
    sharegpt: round (round)*
    round: request response
    request: "HUMAN: " user_string
    response: "GPT: " gpt_string
    user_string: /(?s).+?(?=(?:GPT: |HUMAN: |$))/
    gpt_string: /(?s).+?(?=(?:GPT: |HUMAN: |$))/
Do NOT change the content of the conversation. 
For example: If the input conversation is: "How are you? I'm fine." You should reformat it as "HUMAN: How are you? GPT: I'm fine."
\end{lstlisting}

\subsection{Further Analysis on CFG Reformat as Self-debugging}

We compare the performance of vanilla \PE{} and \PE{} with CFG reformat on all evaluation items in \cref{fig:method_reformat}. %
Self-debugging after voting directly reformats voted samples, which are taken as output or utilized as seeds in the next PE iteration without further selection, resulting in higher \cfg{} while lower performance on semantic and statistic properties.

\begin{figure}[htbp]
    \centering
    \includegraphics[width=0.7\linewidth]{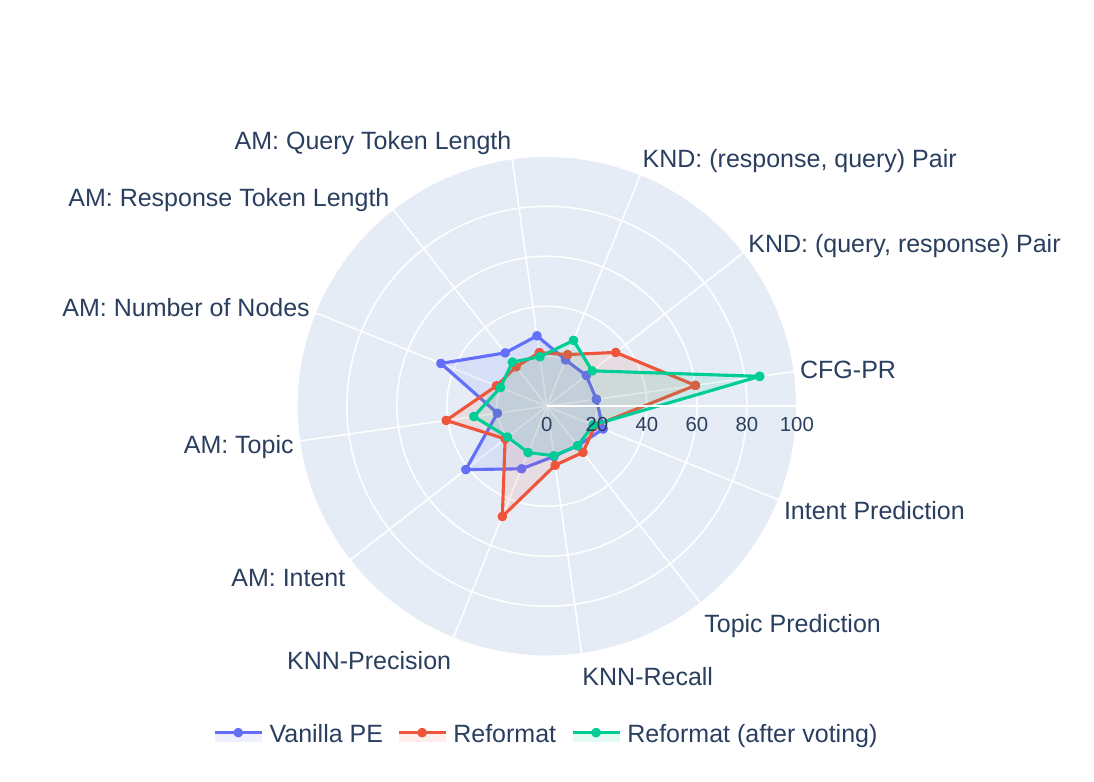}
    \caption{Performance of \PE{} with CFG Reformat on ShareGPT with $\epsilon=4$}
    \label{fig:method_reformat}
\end{figure}

\subsection{Improving Node Dependency (\kns{}): Fix Format Token in Variation API}

Key node dependency (i.e., \kns{}) is an important semantic metric that measures the similarity of the node pair dependencies between the private and synthetic datasets. To improve \kns{}, we fix the format tokens during blank-filling in variation API. Since nodes are recognized and separated by format tokens in textual datasets, fixing the format tokens ensures that multiple nodes will not be mistakenly merged into one and thus helps to remain the original semantic meaning of each node, and therefore the node semantic dependencies. We compare the performance of vanilla \PE{} and \PE{} with fixed format token on \kns{} on (query, response) and (response, query) pairs and \cfg{} in \cref{fig:case_study_fix_token}, where we consider two variants of
our method: fix all the format tokens (shown as Fixed Token) and randomly fix 65\% of the format tokens (shown as Fixed Selected Token). We can observe that our methods achieve better semantic performance on \kns{} compared to vanilla \PE{}, and Fixed Token outperforms since it keeps more node structures than Fixed Selected Token. Additionally, as fixing format tokens avoids node merging, it also improves the structural validity, i.e., \cfg{}.

\begin{figure}[htbp]
    \centering
\begin{subfigure}{0.55\textwidth}
         \centering
    \includegraphics[width=1\linewidth]{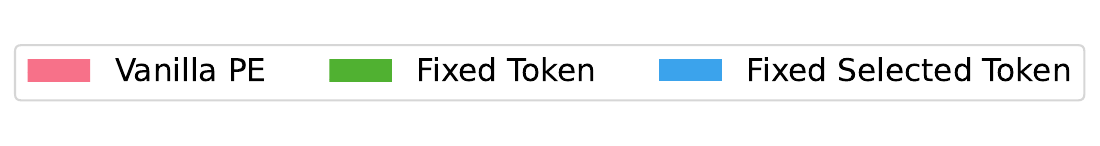}
\end{subfigure}

\begin{subfigure}{0.31\textwidth}
         \centering
    \includegraphics[width=1\linewidth]{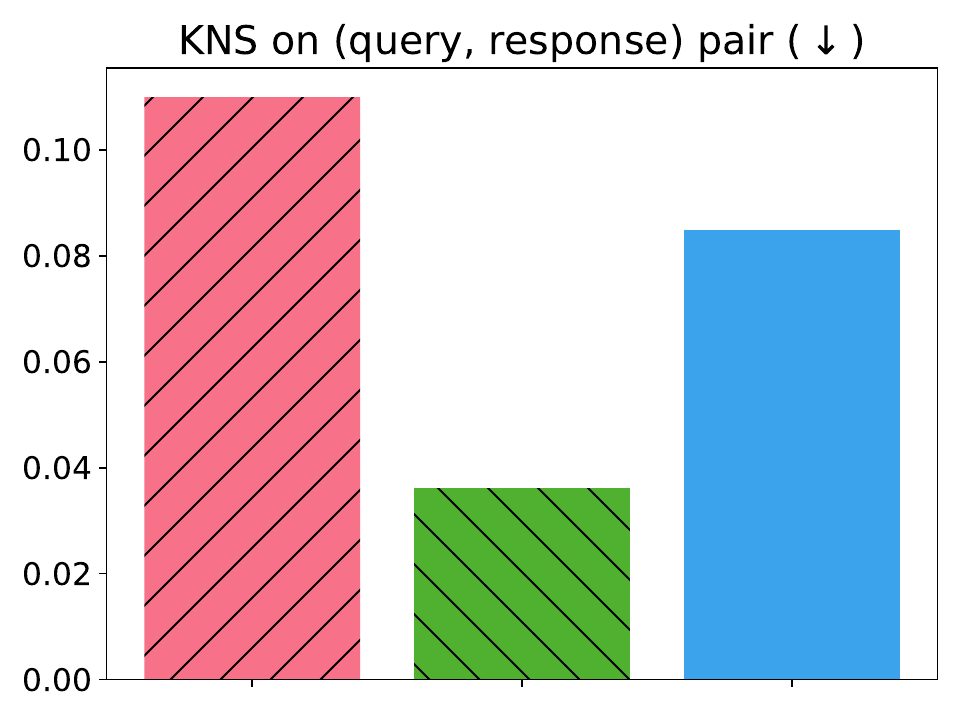}
\end{subfigure}
\begin{subfigure}{0.31\textwidth}
         \centering
    \includegraphics[width=1\linewidth]{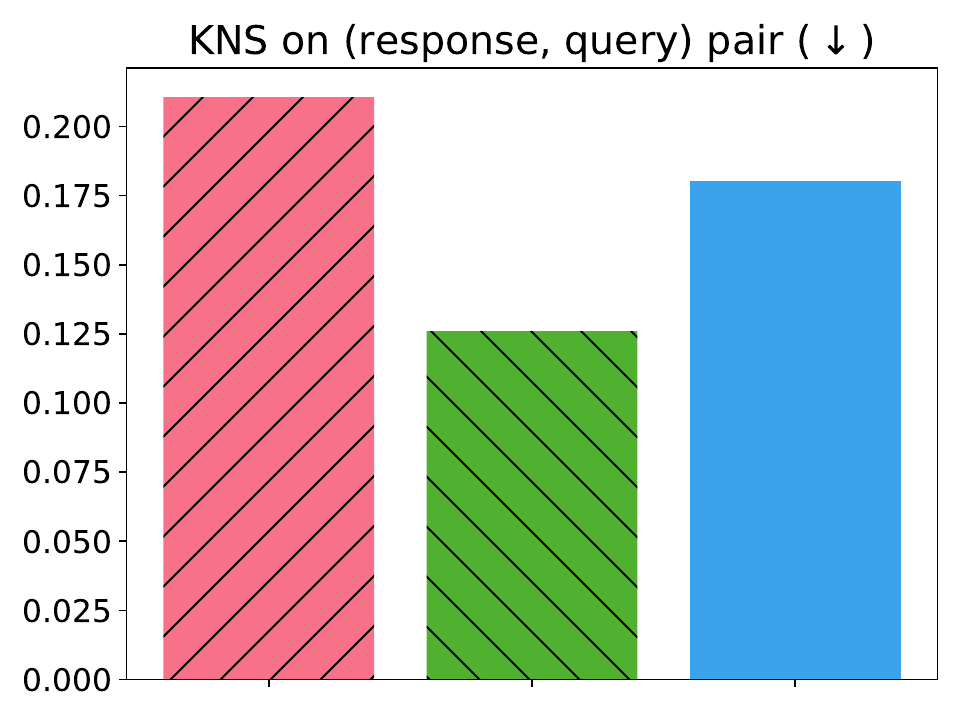}
\end{subfigure}
\begin{subfigure}{0.31\textwidth}
         \centering
    \includegraphics[width=1\linewidth]{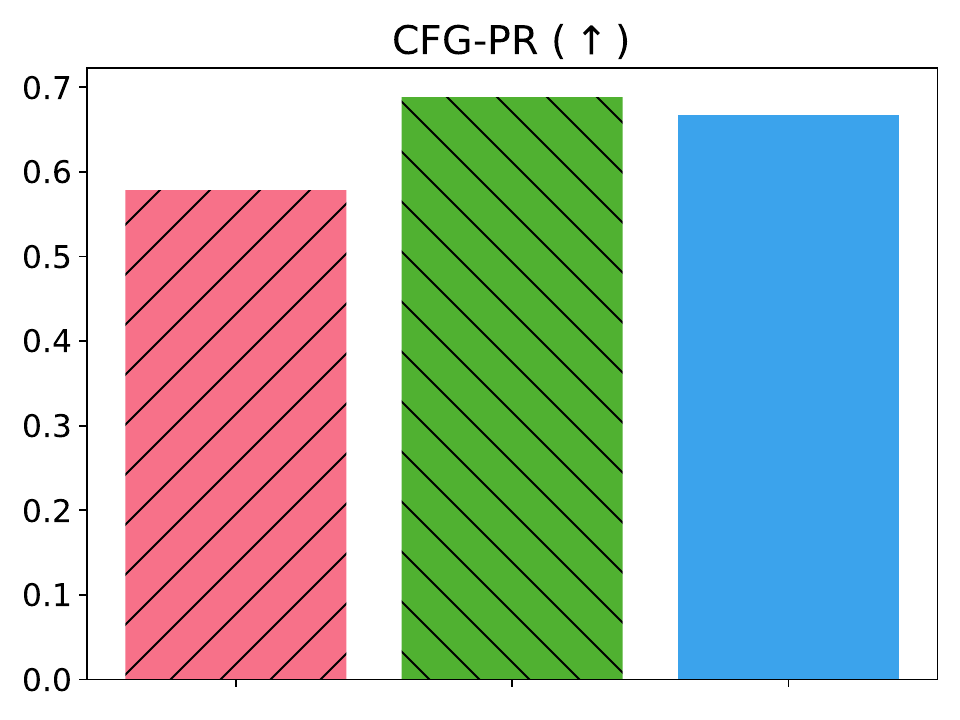}
\end{subfigure}
\caption{Performance of Vanilla PE and \PE{} with fix token on \cfg{} and \kns{}}
\label{fig:case_study_fix_token}
\end{figure}

We then compare the performance of vanilla \PE{} and our methods on all metrics in \cref{fig:method_fix_token}. %
Since Fixed Token fixes all format tokens, the blank-filling process becomes less flexible, e.g., the number of nodes after blank-filling will never decrease, which is ensured by existing format tokens. Therefore, its performance in most statistic properties is worse than that of vanilla \PE{} and Fixed Selected Token.

\begin{figure}[htbp]
    \centering
    \includegraphics[width=0.7\linewidth]{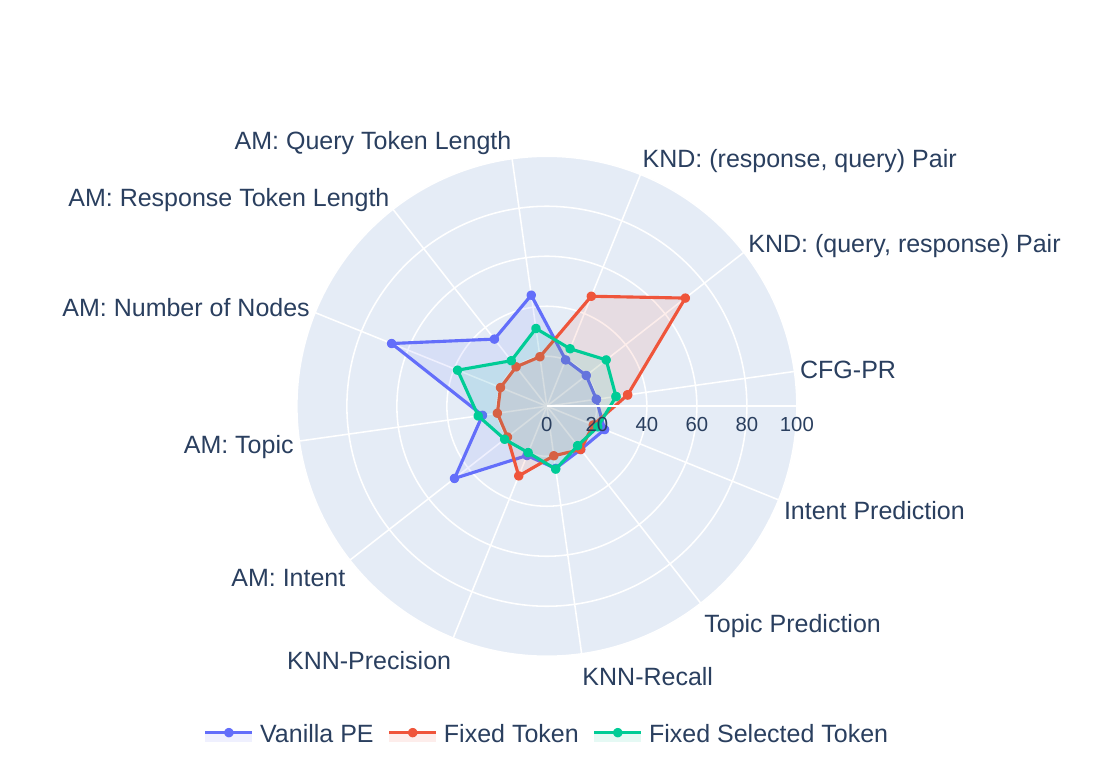}
    \caption{Performance of \PE{} with Fix Format Token on ShareGPT with $\epsilon=4$}
    \label{fig:method_fix_token}
\end{figure}

\subsection{Further Analysis on Node extraction \& Auto-generation}

To further examine the semantic diversity of the dataset, we adopt another metric Type to Token Ratio (TTR) \cite{richards1987type} to provide auxiliary information.
TTR measures diversity in the tokens used in the dataset by dividing the number of unique tokens 
by the total number of tokens in the dataset. A higher TTR suggests a more diverse vocabulary.
\cref{fig:ttr} shows that Extract Query has a higher TTR than vanilla \PE{}.

\begin{figure}[htbp]
    \centering
    \includegraphics[width=0.35\linewidth]{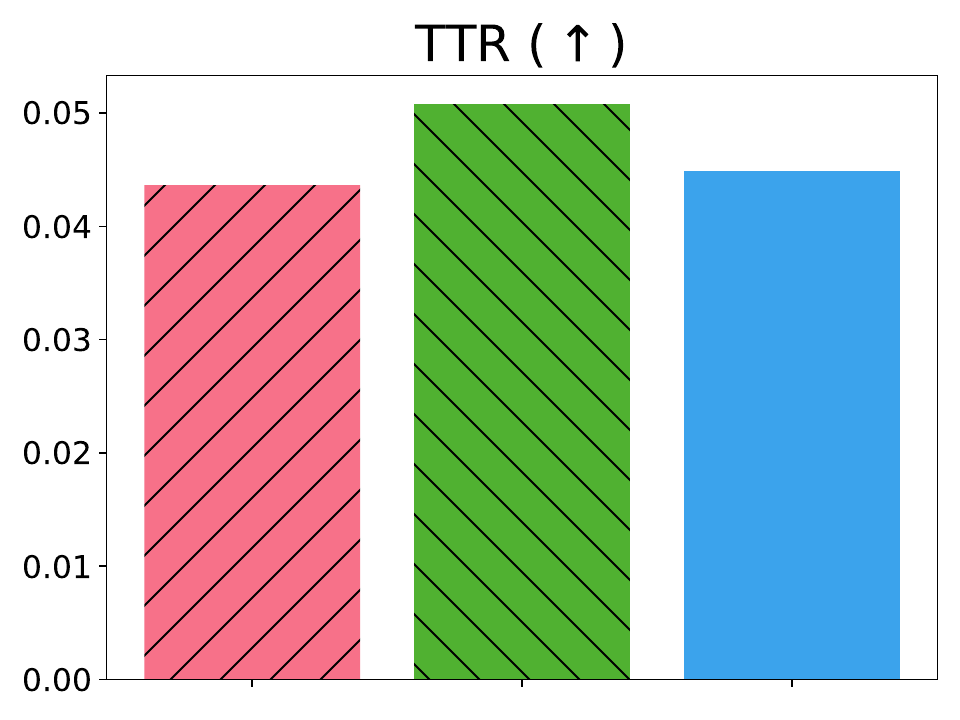}
    \caption{Performance of vanilla PE and \PE{} with node extraction on Type to Token Ratio (TTR).}
    \label{fig:ttr}
\end{figure}

To illustrate the semantic quality and diversity of the synthetic dataset, we then focus on the embeddings of the generated sample, and draw them in a 2-dimensional plot after principal component analysis (PCA). As shown in \cref{fig:case_study_emb_dis_vanilla,fig:case_study_emb_dis_extract}, the embeddings of vanilla \PE{} and \PE{} with query node extraction (blue dots) are drawn together with the embeddings of private data (yellow dots). We can easily observe that the embeddings of \PE{} with query node extraction have more overlaps with the private data embeddings, indicating a higher sample semantic quality and diversity.

\begin{figure}[htbp]
    \centering
\begin{subfigure}{0.4\textwidth}
         \centering
    \includegraphics[width=1\linewidth]{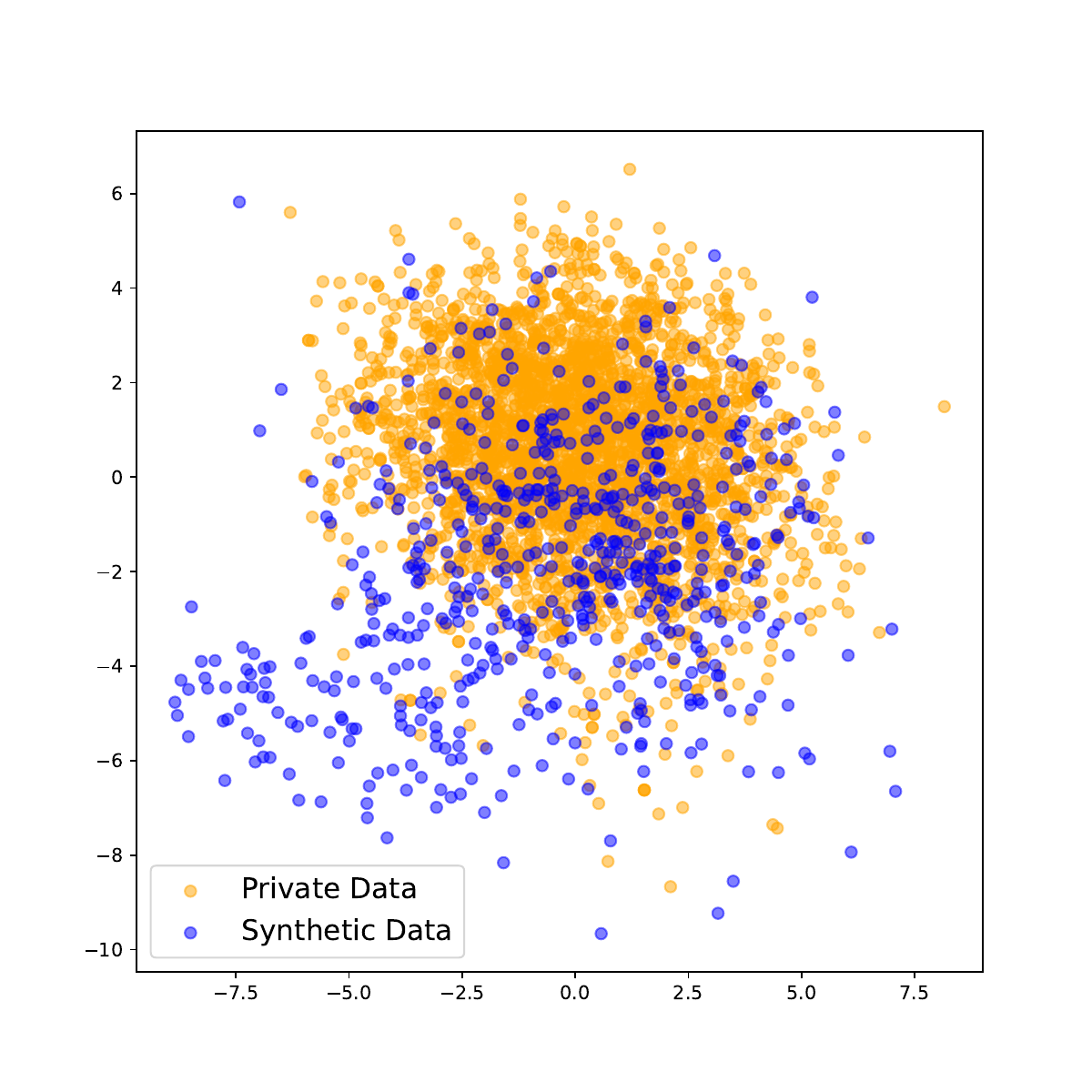}
    \caption{Embeddings of Vanilla \PE{}}
    \label{fig:case_study_emb_dis_vanilla}
\end{subfigure}
\begin{subfigure}{0.4\textwidth}
         \centering
    \includegraphics[width=1\linewidth]{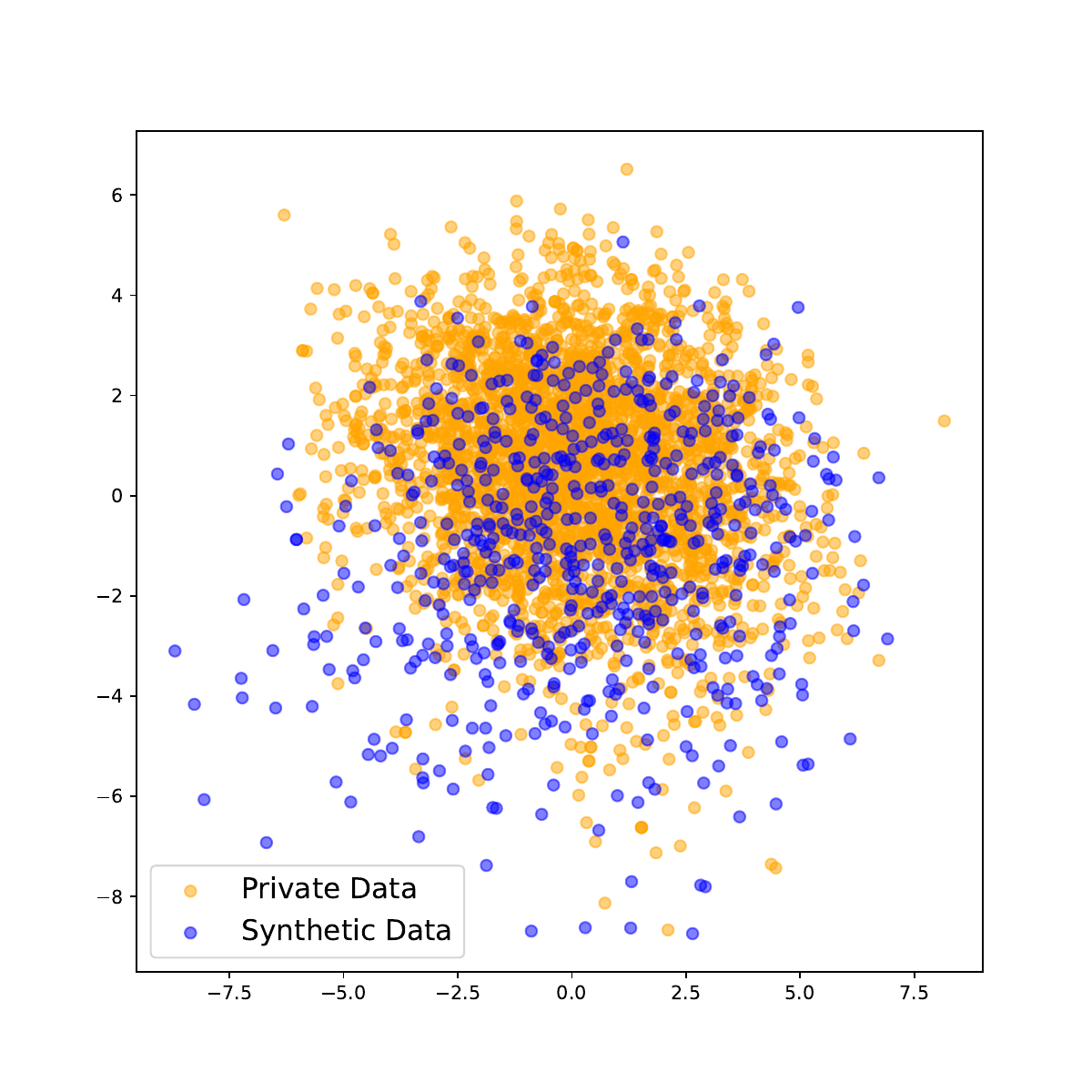}
    \caption{Embeddings of \PE{} with node extraction}
    \label{fig:case_study_emb_dis_extract}
\end{subfigure}
\caption{Embedding distributions of Vanilla \PE{} and \PE{} with node extraction.}
\label{fig:case_study_extract_emb_dis}
\end{figure}

We them compare the performance of vanilla \PE{} and \PE{} with node extraction on all metrics in \cref{fig:method_extract}, where we consider several variants of our method: extract all query nodes and auto-generate all response nodes (shown as Extract Query); combination of query node extraction, reformat before voting, and fix format token (Extract Query \& Reformat \& Fixed Token); combination of query node extraction, reformat before voting, and fix 65\% format token (Extract Query \& Reformat \& Fixed Selected Token); combination of response node extraction, reformat before voting, and fix 65\% format token (Extract Response \& Reformat \& Fixed Selected Token). We can observe that (1) Extract Query outperforms vanilla \PE{} across most statistic properties, \cfg{}, and semantic properties including \precision{}, \recall{}, and \kns{} on (response, query) pair. (2) Extract Query \& Reformat \& Fixed Selected Token outperforms or achieves similar performance to other node extraction variants on \cfg{}, most statistic and semantic metrics. This indicates that the combination of reformat and fix selected format tokens to node extraction improves \cfg{} and structural semantic performance without degrading statistic performance. (3) Extracting query nodes outperforms extracting response nodes, indicating that the type of nodes extracted significantly influences the synthetic data performance.

\begin{figure}[htbp]
    \centering
    \includegraphics[width=0.66\linewidth]{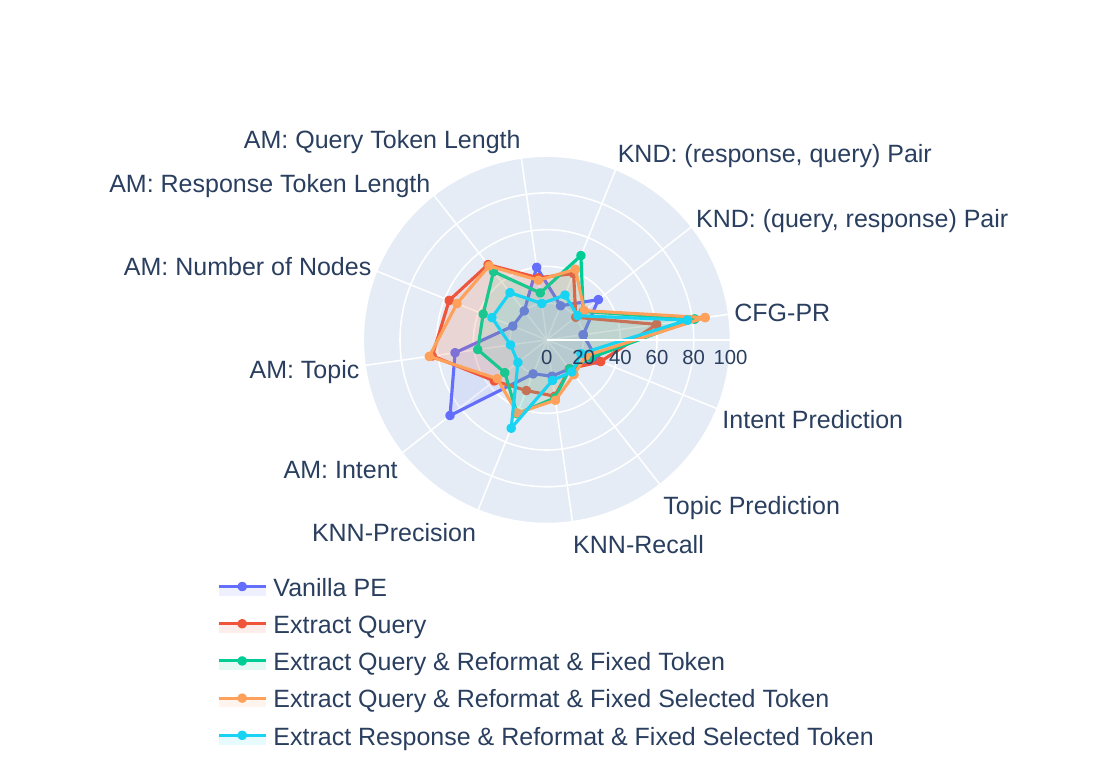}
    \caption{Performance of \PE{} with Node Extraction on ShareGPT with $\epsilon=4$}
    \label{fig:method_extract}
\end{figure}

\subsection{Performance Comparison between Different Methods}

We compare the performance of our proposed methods and some combinations of them according to our benchmark. Specifically, in \cref{fig:case_study_mix}, we illustrate the performance of vanilla \PE{}; \PE{} with CFG reformat; \PE{} with fixed format token; combination of CFG reformat and fix format token (Fixed Token \& Reformat); and combination of CFG reformat, fix partial format token, and query node extraction (Extract Query \& Reformat \& Fixed Selected Token). As we can observe, Extract Query \& Reformat \& Fixed Selected Token outperforms on structural validity \cfg{}, semantic properties \precision{} and \recall{}, and statistic properties \knd{} on conversation round and response token length; while Fixed Token \& Reformat outperforms mainly on semantic properties \kns{} on (query, response) and (response, query) pair. As different methods focus on different aspects of the synthetic data, users can choose the method according to their practical needs. The algorithm design and analysis based on our benchmark also pave the way to propose a method that outperforms on all evaluation metrics, which we leave as a future work.

\begin{figure}[htbp]
    \centering
    \includegraphics[width=0.66\linewidth]{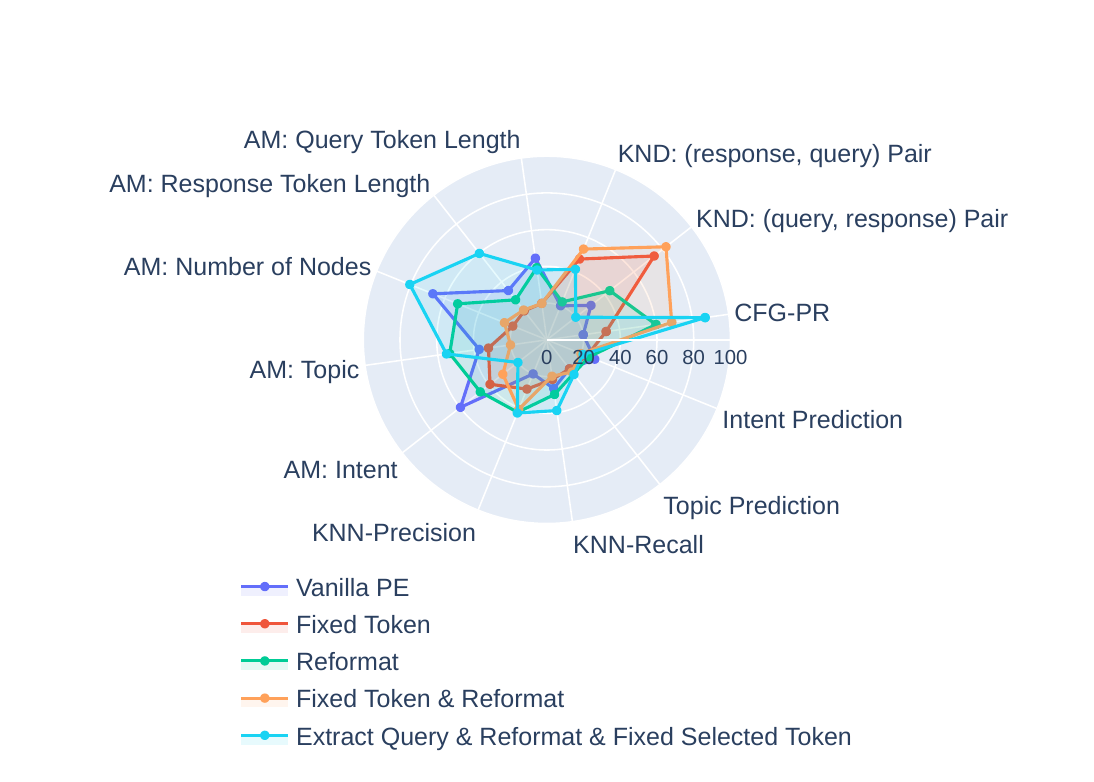}
    \caption{Performance of Different Methods on ShareGPT with $\epsilon=4$}
    \label{fig:case_study_mix}
\end{figure}

\end{document}